\documentclass[twoside]{article}

\usepackage[accepted]{aistats2026}
\usepackage{multirow}
\usepackage{url} 
\usepackage{graphicx}
 \usepackage{natbib} 
 \usepackage{xcolor}
\usepackage{booktabs}
\usepackage{wrapfig}  
\usepackage{subfigure}

\usepackage{amssymb} 

\begin{document}

%

%

\twocolumn[

\aistatstitle{Spectral Text Fusion: A Frequency-Aware Approach to Multimodal Time-Series Forecasting}

\aistatsauthor{ Huu Hiep Nguyen \And Minh Hoang Nguyen \And Dung Nguyen \And Hung Le}

\aistatsaddress{\textit{Applied Artificial Intelligence Initiative
} \\
\textit{Deakin University}\\
Geelong, Australia} ]

\begin{abstract}

Multimodal time series forecasting is crucial in real-world applications, where decisions depend on both numerical data and contextual signals. The core challenge is to effectively combine temporal numerical patterns with the context embedded in other modalities, such as text. While most existing methods align textual features with time-series patterns one step at a time, they neglect the multiscale temporal influences of contextual information such as time-series cycles and dynamic shifts. This mismatch between local alignment and global textual context can be addressed by spectral decomposition, which separates time series into frequency components capturing both short-term changes and long-term trends. In this paper, we propose SpecTF, a simple yet effective framework that integrates the effect of textual data on time series in the frequency domain. Our method extracts textual embeddings, projects them into the frequency domain, and fuses them with the time series' spectral components using a lightweight cross-attention mechanism. This adaptively reweights frequency bands based on textual relevance before mapping the results back to the temporal domain for predictions. Experimental results demonstrate that SpecTF significantly outperforms state-of-the-art models across diverse multi-modal time series datasets while utilizing considerably fewer parameters. Code is available at \url{https://github.com/hiepnh137/SpecTF}.
\end{abstract}

\section{Introduction}
\label{section:intro}

Time series forecasting plays a critical role across numerous domains, including finance \citep{DBLP:journals/financial_time_series} \citep{mondal2014study}, healthcare \citep{zhang2024trajectory} \citep{kaushik2020ai}, energy \citep{deb2017review} \citep{8489399}, material science \citep{le2025accelerating}, and economics \citep{RePEc:ucp:jnlbus:v:39:y:1965:p:139} \citep{franses1998time}, where accurate predictions enable informed decision-making. While traditional forecasting approaches rely solely on historical numerical data, real-world time series are often influenced by external contextual information available as text-news articles, policy documents, social media, and domain-specific reports. This textual information frequently provides crucial insights into factors driving temporal dynamics that pure numerical analysis cannot capture \citep{liu2025timeseriesanalysisbenefit}.

Although recent efforts have been made to incorporate multi-modality into time series forecasting, existing approaches remain limited. MM-TSFlib \citep{liu2024timemmd} employs a late-fusion architecture that views the whole text sequence as a vector by a separate pipeline with time series before combining outputs with a simple linear weighting mechanism, neglecting the cross-modal interaction. Meanwhile, TaTS \citep{DBLP:journals/corr/abs-2502-08942} relies on the text-periodicity assumption to align text with time series by treating time series-paired texts as auxiliary variables of the time series itself. We argue that they still cannot exploit the textual information well, which frequently encodes contextual signals that simultaneously influence multiple temporal scales. For instance, a central bank policy announcement may trigger immediate market volatility (high-frequency components) while also initiating a prolonged economic trend (low-frequency components). Similarly, news of supply chain disruptions might cause both short-term price spikes and longer-term inventory adjustments.


\begin{figure}[t]
    \centering
    \includegraphics[width=\linewidth]{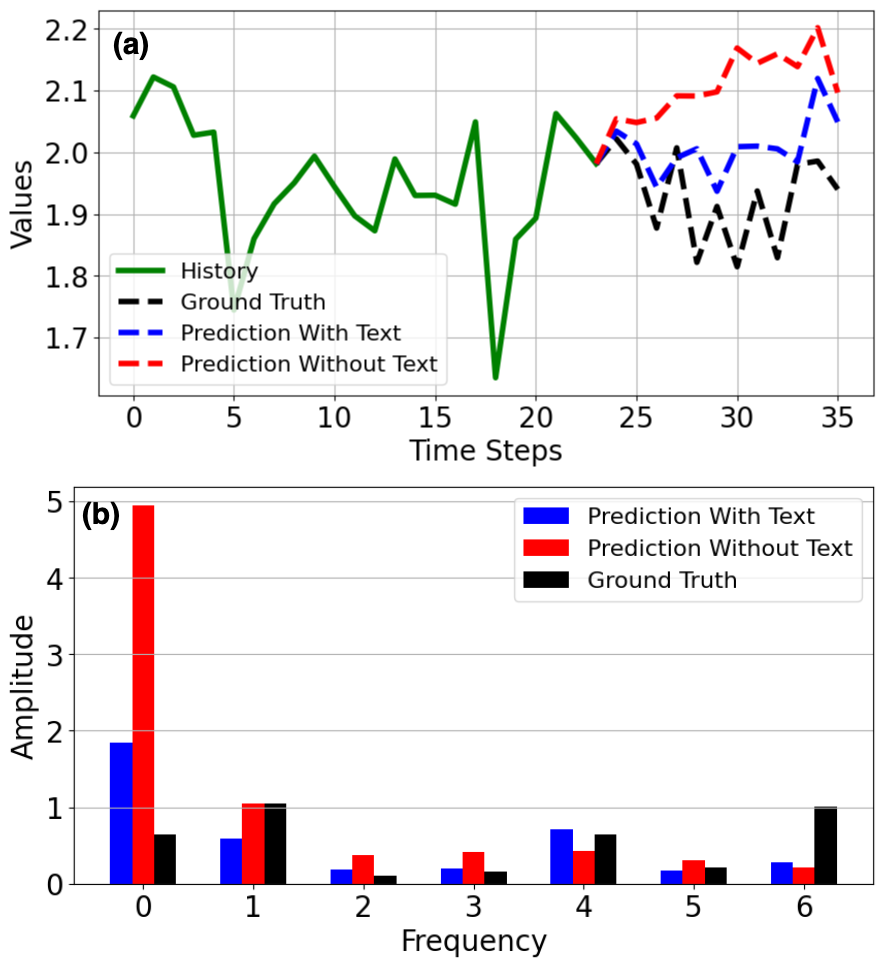}
    \caption{An example of SpecTF's predictions in Agriculture data. We provide both time-domain comparison (a) and frequency-domain comparison (b) in the predictions when incorporating text, compared to when it is not used.}
    \label{fig:analysis_example}
\end{figure}

To overcome the above problems, we explore a novel direction to model the multiple temporal scale influence through the spectral domain, where the influence at high-frequency components represents short-term volatility and the influence at low-frequency components captures prolonged trends. This spectral perspective enables a more nuanced integration of textual context with time series dynamics at their appropriate temporal scales while maintaining parameter efficiency through energy compaction \citep{yi2023frequencydomain}, as real-world time series often concentrate energy in fewer frequency components. 

Specifically, we propose Spectral Text Fusion (SpecTF), a novel framework that models the effect of textual data on time series in the frequency domain. SpecTF decomposes time series into spectral components and explicitly models how textual information modulates distinct frequency bands. As illustrated in Figure~\ref{fig:analysis_example}, our approach demonstrates superior forecasting performance when incorporating text, compared to when it is not used in the Agriculture dataset. To visualize in the spectral domain, we first normalize predictions and the ground truth by the historical mean and variance before using the Fourier Transform \citep{sundararajan2001discrete} to convert to the frequency domain.  The time-domain visualization (Figure~\ref{fig:analysis_example}a) shows that predictions incorporating textual information (blue dashed line) align more closely with the ground truth while the frequency-domain visualization (Figure~\ref{fig:analysis_example}b) reveals that these predictions more accurately match the ground truth across most frequency components, particularly in the 0th, 2nd, 3rd, 4th, 5th, and 6th frequency bands.

The proposed SpecTF framework consists of five components that process multimodal data through the frequency domain (see Figure~\ref{fig:pipeline}). The Time Series Embedding module represents frequency components of the time series by complex vectors. Concurrently, the Text Embedding module processes textual data using a pretrained language model and projects the resulting embeddings into a complex frequency space. The heart of our approach lies in the Frequency Cross-Modality Fusion module (FreqCMF), which implements frequency-domain cross-modal fusion through a frequency attention mechanism that adaptively reweights spectral components based on the relevant textual information. Finally, the Forecaster maps the fused representations to prediction frequency representations before the Projection module converts results back to the temporal domain using the inverse Fourier Transform.


Our contributions can be summarized as follows:

\begin{itemize}
    \item We pioneer a frequency-based approach for multimodal time series that explicitly models the multiscale temporal influence of text on time series by fusing textual and time series data in the spectral domain.
    \item We develop the Frequency Cross-Modality Fusion mechanism (FreqCMF) that uses attention to query relevant textual features from frequency components, enabling adaptive integration of text information based on frequency-specific relevance.
    \item We demonstrate through comprehensive experiments on the Time-MMD benchmark that our method not only surpasses existing multimodal forecasting approaches but also outperforms state-of-the-art time series models adapted for multimodal settings, across diverse domains, achieving state-of-the-art performance while using significantly fewer parameters.

\end{itemize}

\section{Related Works}
\label{section:related_work}
\textbf{Unimodal time series forecasting.} Traditional time series forecasting relies on statistical models to capture temporal patterns. ARIMA \citep{math10213988} \citep{mondal2014study} (seasonal variant) models linear dependencies using autoregressive and moving average terms. Modern architectures leverage self-attention to model temporal dependencies at scale. Temporal-based models like PatchTST \citep{nie2023a} segment time series into patches for local feature extraction, while iTransformer \citep{liu2024itransformer} inverts token-channel relationships to improve noise robustness. TimeMixer \citep{wang2023timemixer}  introduces a multiscale mixing architecture that decomposes time series into seasonal and trend components across multiple sampling scales, applying bottom-up seasonal mixing and top-down trend mixing to disentangle intricate temporal variations. Despite their impressive performance, these methods cannot exploit the external information from other modalities like text, which carries rich semantic information for time-series analysis \citep{liu2024timemmd,ijcai2024p921}. 

\textbf{Frequency-domain approaches in time series.}
The application of frequency-domain techniques to time series forecasting has gained renewed interest due to their ability to isolate multiscale temporal patterns through spectral decomposition \citep{10.1145/3711896.3736571}. AutoFormer \citep{wu2021autoformer} proposes the auto-correlation mechanism using Fourier transforms, and FEDformer \citep{zhou2022fedformer} employs Fourier-enhanced attention for long-term forecasting. FreTS \citep{yi2023frequencydomain} further optimizes frequency-domain learning by redesigning MLPs to exploit global dependencies and energy concentration in spectral components. FITS \citep{xu2024fits} introduces a lightweight framework using complex-valued neural networks for frequency-domain interpolation. However, these approaches focus exclusively on unimodal numerical series, leaving unresolved the critical challenge of modeling the interaction between auxiliary modalities (like text), and time series data in spectral space. Our SpecTF addresses this gap as the first framework to enable frequency-aware multimodal time series forecasting through an attention-based cross-modality fusion mechanism to model the influence of textual data in different frequency components.

\textbf{Multimodal time series forecasting with textual data.}
Multimodal approaches address the limitations of unimodal methods by integrating complementary data sources \citep{10.1145/3711896.3736567}. 
Text2Freq \citep{lo2024textfreq} pioneers frequency-domain alignment but it utilizes textual data that essentially serves as a description or representation of the time series itself; the text is closely tied to the numerical data and does not provide additional or external contextual information.  MM-TSFlib \citep{liu2024timemmd} linearly combines the predictions from text and time series from separate models, neglecting the cross-modal interaction. TaTS \citep{DBLP:journals/corr/abs-2502-08942} projects text embeddings as auxiliary time series variables but conflates their influence across frequencies.  TimeXer \citep{wang2024timexer} represents a significant advancement by empowering canonical Transformers, employing a global token mechanism to bridge information between exogenous and endogenous variables by using the attention mechanism at the temporal domain. While effective for general information transfer, this approach treats the influence of exogenous variables uniformly across all temporal dynamics. Compared to TaTS or TimeXer, which use time-domain fusion, our SpecTF, using cross-attention in the frequency domain, explicitly models the influence of textual data at multiple temporal scales through different frequency components. 

\section{Background}
\label{section:background}
\subsection{Multimodal Time Series Forecasting}

In the context of multimodal time series forecasting, the primary goal is to predict future values of a time series while utilizing additional modalities that are temporally paired with the time series. This paper focuses specifically on text-enhanced forecasting.
In particular, given a collection of multivariate time series samples and textual data with a lookback window \( L \), the number of variables \(N\), we denote the input as pairs \((x_1, s_1), \ldots, (x_L, s_L)\), where \( X^h = \left[x_1, \ldots, x_L\right] \in \mathbb{R}^{N \times L} \) is the numerical data and \( S^h = \left[s_1, \ldots, s_L \right]\) denotes the associated textual data. The objective of multimodal time series forecasting is to predict future numerical values \( Y = \left[x_{L+1}, \ldots, x_{L+H}\right] \in \mathbb{R}^{N \times H} \), where \( H \) is the forecast horizon.

\subsection{Frequency Domain in Time Series}

Frequency domain analysis of time series data provides a powerful framework for decomposing temporal signals into constituent periodic components. This approach offers significant advantages, particularly for capturing multi-scale patterns and global dependencies. The Discrete Fourier Transform (DFT) \citep{sundararajan2001discrete} serves as the fundamental mathematical tool for this transformation, mapping a discrete time series $X=\{x_t\}_{t=0}^{n-1}$ into frequency domain $\hat{X}=\{\mathcal{X}_k\}_{k=0}^{n-1}$:

\begin{equation}
\centering
    \begin{aligned}
        \hat{X} &= \{\mathcal{X}_k\}_{k=0}^{n-1} = \text{DFT}(X), \\
        \mathcal{X}_k &= \sum^{n-1}_{t=0}x_t\exp{-2\pi j \frac{kt}{n}}
    \end{aligned}
\end{equation}

where $\mathcal{X}_k$ represents the complex amplitude and phase of the frequency component at $\omega_k = \frac{2\pi k}{n}$.
The inverse Discrete Fourier Transform (iDFT) enables the reconstruction of the original time series from its frequency representation:

\begin{equation}
\centering
    \begin{aligned}
        X &= \{x_t\}_{t=0}^{n-1} = \text{iDFT}(\hat{X}), \\
        x_t &= \frac{1}{n}\sum^{n-1}_{k=0}\mathcal{X}_k \exp{2\pi j \frac{kt}{n}}.
    \end{aligned}
\end{equation}

The bidirectional mapping formed by DFT and iDFT provides a framework for manipulating specific frequency bands, enabling precise control over different temporal scales, from high-frequency components (representing short-term fluctuations) to low-frequency components (capturing long-term trends). 

The Fast Fourier Transform (FFT) \citep{5217220} is an efficient algorithm for computing the DFT, reducing computational complexity from $\mathbf{O}(n^2)$ to $\mathbf{O}(n\log n)$, which is crucial for large-scale or real-time signal processing. The Real Fast Fourier Transform (rFFT) \citep{5217220} is a specialized variant of the FFT designed to exploit this redundancy by calculating only the first  $n/2+1$ frequency components, corresponding to the non-redundant positive frequencies. The inverse operation, known as the inverse rFFT (irFFT), reconstructs the original real-valued time series from its reduced set of frequency components.
\begin{equation}
\centering
    \begin{aligned}
        \hat{X} &= \{\mathcal{X}_k\}_{k=0}^{n/2} = \text{rFFT}(X), \\ 
        X &= \{x_t\}_{t=0}^{n-1} = \text{irFFT}(\hat{X}). 
    \end{aligned}
\end{equation}

\subsection{Frequency-domain MLP}
\label{FreMLP}
Frequency-domain MLPs exploit these advantages through architectures redesigned for complex-valued inputs. Unlike time-domain MLPs that suffer from point-wise mapping constraints and information bottlenecks \citep{yi2023frequencydomain}, frequency-domain variants operate on compact spectral features. ~\citet{yi2023frequencydomain} and \citet{xu2024fits} use  FreqMLP, a variant of MLPs with complex parameters, to learn in the frequency domain. Specifically, the FreqMLP can be formulated as follows: $(\hat{Z}\hat{W}+\hat{b}),$
\begin{equation}
    \centering
    \begin{aligned}
        Y = \text{FreqMLP}(\hat{Z}) = \sigma(\hat{Z}\hat{W}+\hat{b}),
    \end{aligned}
\end{equation}
where $\hat{Z}\in \mathbb{C}^{L\times d}$ is the frequency spectrum, $\hat{W}=(W_r+jW_i) \in \mathbb{C}^{d\times d'}$ is the complex number weight matrix with $W_r \in \mathbb{R}^{d\times d'}$ and $W_i \in \mathbb{R}^{d\times d'}$, and $\hat{b} = (b_r+jb_i) \in \mathbb{C}^{d'}$ are the complex number biases with $b_r \in \mathbb{R}^{d'}$ and $b_i \in \mathbb{R}^{d'}$. $Y\in \mathbb{C}^{L\times d'}$ is the output of frequency-domain MLP.

\section{Method}
\label{section:method}
\begin{figure*}[t]
    \centering    \includegraphics[width=\linewidth, trim={0pt 0pt 320pt 0pt}, clip]{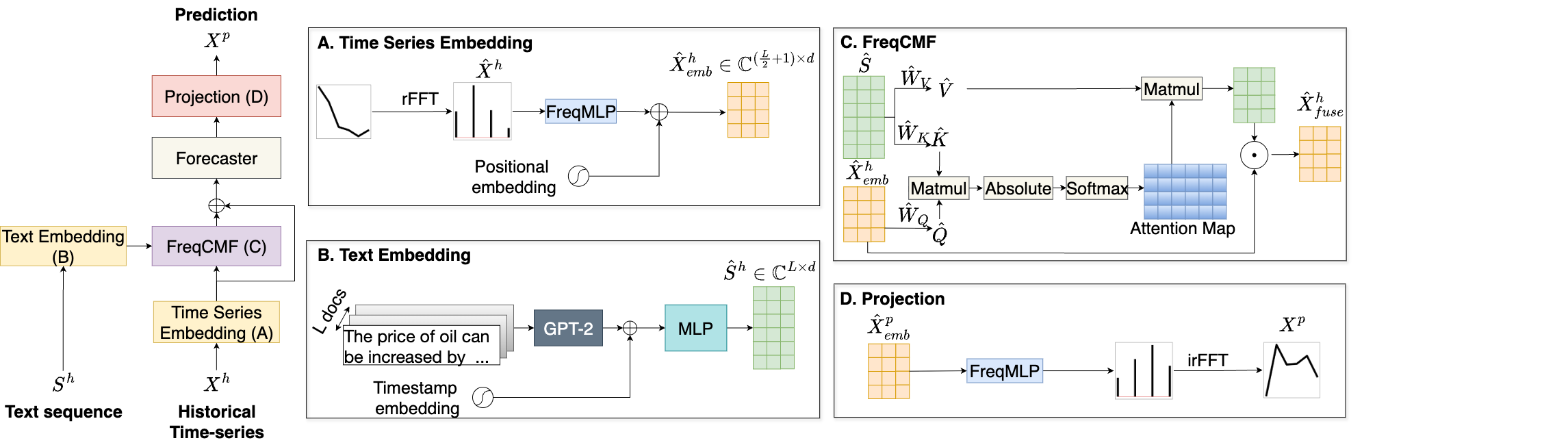}

       \caption{\textbf{An overview of the proposed SpecTF framework.} The time series input is passed through \textbf{Time Series Embedding} to be encoded in the frequency domain, and concurrently, \textbf{Text Embedding} represents each document in the frequency domain. \textbf{Frequency Cross-Modality Fusion (FreqCMF)} integrates textual information into time series frequencies through attention mechanisms and multiplication fusion based on complex multiplication operation $\odot$. \textbf{Forecaster} that maps historical frequency representations to prediction representations in the frequency domain. \textbf{Projection} projects the representations to the original dimension of the spectrum and converts them back to the temporal domain.}
    \label{fig:pipeline}
\end{figure*}

\subsection{SpecTF Pipeline}

Our proposed SpecTF framework, as illustrated in Figure~\ref{fig:pipeline}, consists of five key components designed to effectively model the influence of textual information on time series across multiple frequency bands. We employ the channel-independent strategy \citep{nie2023a} \citep{Zeng_Chen_Zhang_Xu_2023} for our approach, where each variate in the input time series is fed independently into SpecTF.

\textbf{Time Series Embedding.} Given a historical univariate time series $X^h \in \mathbb{R}^{1\times L}$ of length $L$ and associated textual information $S^h$, we first transform $X^h$ into the frequency domain using rFFT, resulting in complex-valued spectral components. Then, we use a frequency-domain MLP (FreqMLP) as mentioned in Section~\ref{FreMLP} as an embedding layer to map the spectrum to a higher-dimensional space. Additionally, we incorporate the positional encoding \citep{NIPS2017_3f5ee243} to assign each frequency band (from low to high) a unique high-dimensional signature with particular patterns across embedding dimensions:

\begin{equation}
\centering
    \begin{aligned}
        \hat{X}^h &= \text{rFFT}(X^h), \\
        \hat{X}_{emb}^h &= \text{FreqMLP}(\hat{X}^h) + \text{PoE}(\hat{X}^h)
    \end{aligned}
\end{equation}

where $\hat{X}^h_{emb} \in \mathbb{C}^{(L/2+1)\times d}$, $d$ is the dimension of frequency vectors, PoE($\cdot$) is the positional encoding of a sequence.  

\textbf{Text Embedding.} Concurrently, the textual data $S^h$ is processed through our Text Embedding block, which projects the input text into a complex embedding space $\hat{S}^h = \text{TextEmbedding}(S^h)$ with  $\hat{S}^h \in \mathbb{C}^{L \times d}$. 
Here, we employ a pre-trained language model to extract contextual embeddings, followed by an MLP layer that maps these embeddings into the complex embedding space (see details in Section~\ref{subsec:textembd}). 

\textbf{FreqCMF.} Our framework's key innovation lies in the fusion approach implemented via  Frequency Cross-Modality Fusion (FreqCMF) block. It incorporates textual context into historical frequency components $\hat{X}^h_{fuse} = \text{FreqCMF}(\hat{X}^h_{emb}, \hat{S}^h)$, where $\hat{X}^h_{fuse} \in \mathbb{C}^{(L/2+1)\times d}$, The inner operations of FreqCMF will be given in Section~\ref{subsec:freqcmf}. This fusion block enables our model to leverage textual information for understanding historical patterns.

\textbf{Forecaster.} Then, we employ a Forecaster model that maps these historical frequency components to future ones
$\hat{X}^{p}_{emb} \in \mathbb{C}^{(H/2+1)\times d}$.
Here, the Forecaster is implemented as a FreqMLP layer:
\[
\hat{X}^{p}_{emb}
=
\operatorname{FreqMLP}\!\bigl((\hat{X}^{h}_{fuse})^{\top}\bigr)^{\top}.
\]

\textbf{Projection.} After that, we use a Projection layer to map the embedding of prediction frequency to the original dimension of the spectrum before converting it back to the temporal domain $X^p$ through the  inverse Real Fast Fourier Transform (irFFT):

\begin{equation}
\centering
    \begin{aligned}
        \hat{X}^p &= \text{FreqMLP}(\hat{X}^p_{emb}), \\
        X^p &= \text{irFFT}(\hat{X}^p)
    \end{aligned}
\end{equation}
where $X^p \in \mathbb{R}^{1\times H}$ is the prediction of SpecTF based on the input $X^h$. By operating primarily in the frequency domain and leveraging complex-valued operations, SpecTF achieves remarkable parameter efficiency while explicitly modeling the multi-scale influence of textual information on time series dynamics.

\subsection{Text Embedding}\label{subsec:textembd}
In this block, textual data are presented in a complex vector space. This process involves three stages: (1) Language Model Encoding, (2) Temporal Alignment and (3) Complex Projection.

\textbf{Language Model Encoding.} We extract contextual embeddings from text using a pre-trained language model such as BERT \citep{devlin-etal-2019-bert} or GPT-2 \citep{radford2019language}: $e_t = \text{LM}(s_t)$
where $e_t \in \mathbb{R}^{L\times d_{LM}}$ is the contextual embedding of the text instance corresponding to the time series at step $t$, $d_{LM}$ is the dimension of the embedding.

\textbf{Temporal Alignment.} To align with time series data, we incorporate temporal information by adding timestamp embeddings \citep{Zhou_Zhang_Peng_Zhang_Li_Xiong_Zhang_2021}: $e^{aligned}_t= e_t + e^{temporal}_t$, where $e^{temporal}_t$ encodes the timestamp at step $t$.


\textbf{Complex Projection.} We map the aligned embeddings to real and imaginary components via separate MLPs:
\begin{equation}
    \centering
    \begin{aligned}
        &Re(\hat{s}_t) = \text{MLP}_{Re}(e^{aligned}_t), \\  
        &Im(\hat{s}_t) = \text{MLP}_{Im}(e^{aligned}_t) 
    \end{aligned}
\end{equation}
where $\hat{s}_t = Re(\hat{s}_t) + j Im(\hat{s}_t) \in \mathbb{C}^{L\times d}$, $Re(\cdot)$ and $Im(\cdot)$ represent the real and imaginary parts of a complex vector. As a result, $\hat{S}^h=\left[\hat{s}_1,\dots,\hat{s}_L\right]$ is the output of the Text Embedding module, which is resided in the same complex frequency space as the time series, enabling direct modulation of the spectral component.

\subsection{Frequency Cross-Modality Fusion (FreqCMF)}\label{subsec:freqcmf}

\begin{table*}[t]
\renewcommand{\arraystretch}{1.5}  
\centering
\caption{Forecasting results on Time-MMD benchmark \citep{liu2024timemmd}. The results are averaged across three different seeds and all prediction lengths. The best numbers in each row are shown in \textbf{\textcolor{red}{bold}}, and the second-best numbers are \textcolor{blue}{\underline{underlined}}. The last column denotes the percentage reduction in MSE and MAE achieved by SpecTF compared to the best-performing baseline. Full results are in Appendix~\ref{section:appendix-forecasting-results}} 
\resizebox{\textwidth}{!}{%
\begin{tabular}{l|cc|cc|cc|cc|cc|cc|cc||cc}
\cline{1-17}
\multirow{2}{*}{\textbf{Models}} & \multicolumn{2}{c|}{\textbf{SpecTF (ours)}} & \multicolumn{2}{c|}{\textbf{TaTS}} & \multicolumn{2}{c|}{\textbf{MM-TSF}} & \multicolumn{2}{c|}{\textbf{TimeXer}} & \multicolumn{2}{c|}{\textbf{FreTS}} & \multicolumn{2}{c|}{\textbf{TimeLLM}} & \multicolumn{2}{c||}{\textbf{ChatTime}} & \multicolumn{2}{c}{\textbf{Promotion}} \\ 
\cline{2-17}

 & \textbf{MSE} & \textbf{MAE} & \textbf{MSE} & \textbf{MAE} & \textbf{MSE} & \textbf{MAE} & \textbf{MSE} & \textbf{MAE} & \textbf{MSE} & \textbf{MAE} & \textbf{MSE} & \textbf{MAE} & \textbf{MSE} & \textbf{MAE} & \textbf{MSE} & \textbf{MAE} \\ 
\cline{1-17}
Agriculture
 & \textbf{\textcolor{red}{0.103}} & \textbf{\textcolor{red}{0.218}} & \textcolor{blue}{\underline{0.112}} & \textcolor{blue}{\underline{0.237}} & 0.123 & 0.263 & 0.132 & 0.263 & 0.120 & 0.249 & 0.157 & 0.301 & 0.305 & 0.345 & 8.35\% & 8.02\%\\ 

\cline{1-17}
Climate
 & \textbf{\textcolor{red}{0.938}} & \textbf{\textcolor{red}{0.762}} & \textcolor{blue}{\underline{1.002}} & \textcolor{blue}{\underline{0.790}} & 1.086 & 0.832 & 1.053 & 0.805 & 1.289 & 0.888 & 1.175 & 0.868 & 1.528 & 1.002 & 6.38\% & 3.54\%\\ 

\cline{1-17}
Economy
& \textcolor{blue}{\underline{0.0085}} & \textcolor{blue}{\underline{0.0780}} & \textbf{\textcolor{red}{0.0083}} & \textbf{\textcolor{red}{0.0770}} & 0.0121 & 0.0883 & 0.0096 & 0.0817 & 0.0152 & 0.0986 & 0.0319 & 0.1438 & 0.0547 & 0.1786 & -2.41\% & -1.30\%\\ 

\cline{1-17}
Energy
& \textbf{\textcolor{red}{0.246}} & \textbf{\textcolor{red}{0.359}} & \textcolor{blue}{\underline{0.264}} & \textcolor{blue}{\underline{0.374}} & 0.270 & 0.393 & 0.288 & 0.400 & 0.299 & 0.401 & 0.264 & 0.369 & 0.270 & 0.393 & 6.82\% & 4.01\%\\ 

\cline{1-17}
Environment
& \textbf{\textcolor{red}{0.259}} & \textbf{\textcolor{red}{0.367}} & \textcolor{blue}{\underline{0.266}} & \textcolor{blue}{\underline{0.370}} & 0.391 & 0.494 & 0.272 & 0.371 & 0.372 & 0.482 & 0.342 & 0.441 &  0.391 & 0.494 & 2.63\% & 0.81\%\\ 

\cline{1-17}
Health
& \textbf{\textcolor{red}{1.276}} & \textbf{\textcolor{red}{0.733}} & \textcolor{blue}{\underline{1.340}} & \textcolor{blue}{\underline{0.763}} & 1.508 & 0.897 & 1.394 & 0.973 & 1.542 & 0.814 & 1.699 & 0.871 & 1.508 & 0.897 & 4.78\% & 3.93\%\\ 

\cline{1-17}
Security
& \textbf{\textcolor{red}{108.411}} & \textbf{\textcolor{red}{4.745}} & 110.600 & 5.069 & 116.003 & 5.572 & \textcolor{blue}{\underline{109.472}} & \textcolor{blue}{\underline{4.870}} & 128.873 & 6.260 & 108.812 & 4.836 & 136.273 & 5.985 & 0.97\% & 2.56\% \\ 

\cline{1-17}
Social Good
& \textbf{\textcolor{red}{0.962}} & \textbf{\textcolor{red}{0.443}} & \textcolor{blue}{\underline{1.012}} & \textcolor{blue}{\underline{0.449}} & 1.183 & 0.497 & 1.002 & 0.478 & 1.086 & 0.520 & 1.084 & 0.573 & 1.183 & 0.497 & 4.94\% & 1.34\%\\ 

\cline{1-17}
Traffic
& \textbf{\textcolor{red}{0.171}} & \textbf{\textcolor{red}{0.210}} & 0.192 & 0.218 & 0.197 & 0.233 & \textcolor{blue}{\underline{0.176}} & \textcolor{blue}{\underline{0.210}} & 0.205 & 0.241 & 0.171 & 0.388 & 0.197 & 0.233 & 2.85\% & 0\%\\ 

\cline{1-17}
\end{tabular}%
}
\label{tab:main_result}

\end{table*}

\begin{table*}[t]
\renewcommand{\arraystretch}{1.5}  
\centering
\caption{Forecasting results on TTC benchmark \citep{kim2024multimodalforecasterjointlypredicting}. The results are averaged across three different seeds and all prediction lengths. The best numbers in each row are shown in \textbf{\textcolor{red}{bold}}, and the second-best numbers are \textcolor{blue}{\underline{underlined}}. The last column denotes the percentage reduction in MSE and MAE achieved by SpecTF compared to the best-performing baseline. Full results are in Appendix~\ref{section:appendix-forecasting-results_ttc}} 
\resizebox{0.9\textwidth}{!}{%
\begin{tabular}{l|cc|cc|cc|cc|cc|cc|cc||cc}
\cline{1-17}
\multirow{2}{*}{\textbf{Models}} & \multicolumn{2}{c|}{\textbf{SpecTF (ours)}} & \multicolumn{2}{c|}{\textbf{TaTS}} & \multicolumn{2}{c|}{\textbf{MM-TSF}} & \multicolumn{2}{c|}{\textbf{TimeXer}} &
\multicolumn{2}{c|}{\textbf{FreTS}} & \multicolumn{2}{c|}{TimeLLM} & \multicolumn{2}{c||}{ChatTime} & \multicolumn{2}{c}{\textbf{Promotion}} \\ 
\cline{2-17}

 & \textbf{MSE} & \textbf{MAE} & \textbf{MSE} & \textbf{MAE} & \textbf{MSE} & \textbf{MAE} & \textbf{MSE} & \textbf{MAE} & \textbf{MSE} & \textbf{MAE} & \textbf{MSE} & \textbf{MAE} & \textbf{MSE} & \textbf{MAE} & \textbf{MSE} & \textbf{MAE} \\ 
\cline{1-17}
Climate
 & \textbf{\textcolor{red}{0.693}} & \textbf{\textcolor{red}{0.605}} & 0.790 & 0.636 & 0.802 & 0.662 & \textcolor{blue}{\underline{0.760}} & \textcolor{blue}{\underline{0.631}} & 0.878 & 0.682 & 0.699 & 0.611 & 1.327 & 0.836 & 8.81\% & 4.12\%\\ 

\cline{1-17}
Medical
 & \textbf{\textcolor{red}{1.298}} & \textbf{\textcolor{red}{0.828}} & 1.313 & 0.953 & 1.436 & 1.072 & \textcolor{blue}{\underline{1.311}} & \textcolor{blue}{\underline{0.880}} & 1.511 & 1.073 & 1.572 & 1.104 & 1.610 & 1.117 & 1.00\% & 5.90\%\\ 

\cline{1-17}
\end{tabular}%
}
\label{tab:main_result_ttc}

\end{table*}

This novel component, the cornerstone of our approach, enables the direct integration of textual information with time series spectral components in the frequency domain. Unlike existing methods that operate in the time domain and struggle to capture text's multi-scale influence, our fusion block explicitly models these interactions through operations in the frequency domain. The fusion process involves Frequency Cross Attention (FCA) and Multiplication Fusion (MF) components.

\textbf{Frequency Cross Attention.} Given frequency representations of time series $\hat{X}^h_{emb}$ and the textual complex representation $\hat{S}^h$, we propose to compute the query, key, and value using FreqMLPs:
\begin{equation}
\centering
    \begin{aligned}
    \hat{Q} &= \text{FreqMLP}_q(\hat{X}^h_{emb}), \\
    \hat{K} &= \text{FreqMLP}_k(\hat{S}^h), \\
    \hat{V} &= \text{FreqMLP}_v(\hat{S}^h)
    \end{aligned}
\end{equation}

where $\hat{Q} \in \mathbb{C}^{ (L/2+1)\times d_k}$, $\hat{K} \in \mathbb{C}^{ L \times d_k} $ and $\hat{V} \in \mathbb{C}^{ L \times d}$. 
Then we compute the attention output:
\begin{equation}
    \centering
    \hat{O} = \text{softmax}\left(\frac{|\hat{Q}\hat{K}^\top|}{\sqrt{d_k}}\right) \cdot \hat{V}
\end{equation}
where $d_k$ is the dimension of query and key, $|\cdot|$ extracts the absolute value of the complex products, ensuring that attention weights reflect the strength of alignment between time series queries and textual keys. This attention mechanism identifies relevant textual information for each frequency component.

\textbf{Multiplication Fusion.} Following attention computation, we apply a complex multiplication operation to incorporate textual influence:
\begin{equation}
    \centering
    \begin{aligned}
    \hat{X}_{fuse} &= \hat{X}_{emb} \odot \hat{O}\\
    & = \left(Re(\hat{X}_{emb})Re(\hat{O})-Im(\hat{X}_{emb})Im(\hat{O})\right) + \\&j\left(Re(\hat{X}_{emb})Im(\hat{O}) + Im(\hat{X}_{emb})Re(\hat{O})\right).
    \end{aligned}
\end{equation}
 This operation in the frequency domain simultaneously modifies both amplitude and phase information, allowing textual context to selectively emphasize or attenuate specific frequency bands while also adjusting their temporal alignment. The multiplication in the frequency domain corresponds to convolution in the time domain \citep{yi2023frequencydomain}, enabling a more nuanced integration of textual influence across multiple temporal scales. We provide the theoretical support in Appendix~\ref{sec:theoretical_analysis}.

\section{Experiments}
\label{section:experiments}
\begin{table*}[t]
\renewcommand{\arraystretch}{1.5}  

\centering
\caption{Ablation study on Agriculture, Health, Environment datasets. \textbf{w/o multiplication fusion} removes multiplication fusion. \textbf{w/o frequency cross attention} replaces cross attention by summation operation. \textbf{w/o textual data} excludes textual data. \textbf{w/o time series data} removes time series data. \textbf{w/ real-imaginary MLP} replace the FreqMLPs by applying MLPs independently to the real and imaginary parts. The best numbers in each column are shown in \textbf{\textcolor{red}{bold}}, and the second-best numbers are \textcolor{blue}{\underline{underlined}}.}
\resizebox{0.7\textwidth}{!}{%
\begin{tabular}{l|cc|cc|cc}
\cline{1-7}
\multirow{2}{*}{\textbf{Models}} & \multicolumn{2}{c|}{\textbf{Agriculture}} & \multicolumn{2}{c|}{\textbf{Health}} & \multicolumn{2}{c}{\textbf{Environment}} \\ 
\cline{2-7}

 & \textbf{MSE} & \textbf{MAE} & \textbf{MSE} & \textbf{MAE} & \textbf{MSE} & \textbf{MAE} \\ 
\cline{1-7}
SpecTF (ours)
 & \textbf{\textcolor{red}{0.103}} & \textbf{\textcolor{red}{0.218}} & \textbf{\textcolor{red}{1.276}} & \textbf{\textcolor{red}{0.733}} & \textbf{\textcolor{red}{0.259}} & \textbf{\textcolor{red}{0.367}} \\ 
\cline{1-7}
w/o multiplication fusion & \textcolor{blue}{\underline{0.104}} & \textcolor{blue}{\underline{0.222}} & \textcolor{blue}{\underline{1.289}} & \textcolor{blue}{\underline{0.752}}  & \textcolor{blue}{\underline{0.263}} & \textcolor{blue}{\underline{0.369}} \\
\cline{1-7}
w/o frequency cross attention & 0.115 & 0.234  & 1.291 & 0.751 & 0.272 & 0.370 \\
\cline{1-7}
w/ real–imaginary MLP & 0.113 & 0.230 & 1.297 & 0.750 & 0.330 &  0.423 \\
\cline{1-7}
w/o textual data & 0.132 & 0.262 & 1.342 & 0.776 & 0.355 & 0.368 \\
\cline{1-7}
w/o time series data & 0.160 & 0.311 & 1.668 & 0.895 & 0.379 & 0.491 \\

\cline{1-7}
\end{tabular}%
}
\label{tab:ablation_study}

\end{table*}

\textbf{Datasets.}
We evaluate SpecTF on two comprehensive multimodal time series benchmarks: the Time-MMD dataset \citep{liu2024timemmd} and the TimeText Corpus (TTC) \citep{kim2024multimodalforecasterjointlypredicting}. Further dataset details are provided in Appendix~\ref{section:appendix-experiment-details}.

\textbf{Baselines.}
We compare our SpecTF with the representative and state-of-the-art models for both unimodal and multimodal time series forecasting. For \textit{unimodal} forecasting, we compare SpecTF against FreTS \citep{yi2023frequencydomain}, a frequency-based method. For \textit{multimodal} time series forecasting, we include MM-TSF \citep{liu2024timemmd} and TaTS \citep{DBLP:journals/corr/abs-2502-08942} with their best variant using iTransformer \citep{liu2024itransformer} backbone for comparison. In addition, we also include TimeXer \citep{wang2024timexer} with text sequence as an exogenous variable by using our Text Embedding block rather than the exogenous embedding layer. For \textit{LLM-based models}, we compare SpecTF against TimeLLM \citep{jin2024timellm} and ChatTime \citep{chattime}.

\textbf{Implementation Details.}
We conduct all experiments on a single NVIDIA Tesla V100. We take MSE (Mean Squared Error) as the loss function and MAE (Mean Absolute Error) as the evaluation metric. GPT2 \citep{radford2019language} is used to embed the text. For additional implementation details, please refer to Appendix~\ref{section:appendix-experiment-details}.

\subsection{Benchmarking Results}

Table~\ref{tab:main_result} presents the comparative results of SpecTF against all baselines across the nine domains from Time-MMD benchmark. For each dataset, we report the results averaged across three different seeds and all prediction lengths.  Overall, SpecTF outperforms both unimodal and multimodal baselines in terms of MSE and MAE metrics on 8 out of 9 domains. When compared to the unimodal baseline FreTS and LLM-based approaches (TimeLLM, ChatTime), SpecTF demonstrates consistent improvements across all domains. This indicates that incorporating textual information in the frequency domain provides valuable additional context for forecasting. In comparison to the runner-up among multimodal approaches (TimeXer, MM-TSF, and TaTS), SpecTF still achieves superior performance in most domains, with the average improvement across 9 datasets of 3.82\%  and 2.25\% in MSE and MAE, respectively. The most significant improvements over multimodal baselines are observed in the Energy, Environment, Health, and Security domains, where textual information often contains signals about regime shifts or policy changes that manifest at different temporal scales. However, SpecTF underperforms on the Economy dataset, where TaTS achieves marginally better results because the textual data contains more noise and shorter-term economic indicators that may be better suited to the time-domain alignment mechanism of TaTS. 

Table~\ref{tab:main_result_ttc} shows SpecTF's performance on TTC benchmark, averaged over prediction lengths. Due to computational constraints, we limit our comparisons to the best-performing baselines from earlier experiments. SpecTF achieves superior results on both datasets: 8.81\% improvements in MSE  on Climate, and 5.90\% improvements in MAE on Medical, demonstrating robustness across diverse multimodal datasets.

\begin{figure*}[t]
    \centering
    \includegraphics[width=\linewidth]{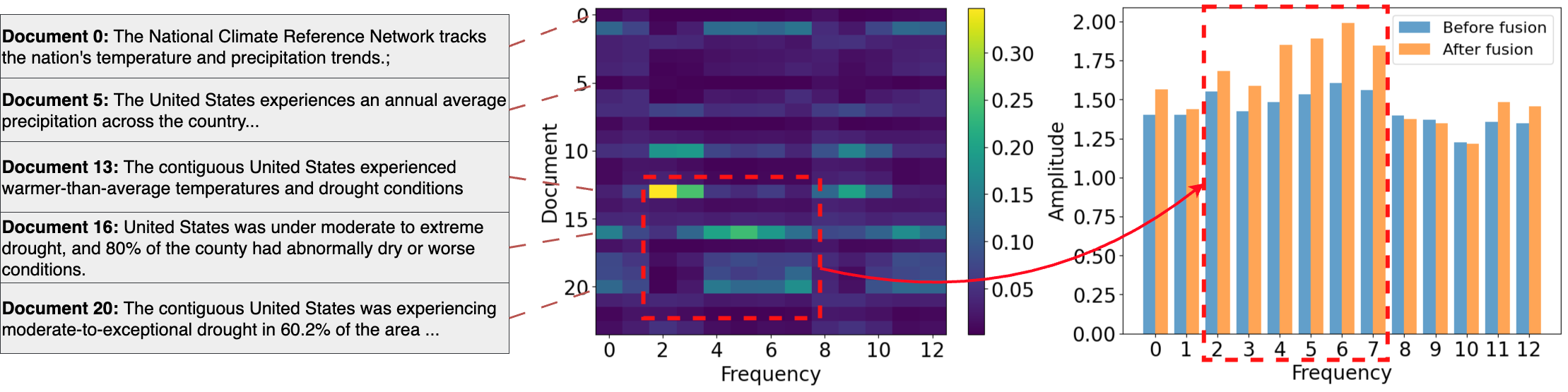}

   \caption{\textbf{Visualization of frequency-text interactions in the Climate dataset.} The attention map (center) reveals document-frequency relationships, with brightness indicating attention strength. Documents 13, 16, and 20 (left panel) show strong attention weights in middle-frequency bands (2-7), highlighted by the red rectangle. The frequency amplitude comparison (right) confirms this relationship: amplitudes after fusion (orange) show significant increases in these middle frequencies compared to the values before fusion (blue), as indicated by the corresponding red rectangle.}
    \label{fig:analysis_attention_map}
\end{figure*}

\subsection{Model Analysis}
\textbf{Ablation study.}
We perform an ablation study spanning three domains (Table~\ref{tab:ablation_study}), comparing SpecTF with four alternative model variants: (1) w/o multiplication fusion, which removes the multiplication fusion; (2) w/o frequency cross attention, substituting cross-attention with direct embedding addition; (3) w/ real–imaginary MLP, replacing FreqMLP with separate MLPs applied independently to real and imaginary components; (4) w/o textual data, excluding all textual inputs to assess modality dependence; and (5) w/o time series data, removing numerical sequences to test text-only forecasting viability.  The results demonstrate the multiplication fusion mechanism’s critical role, with its removal causing performance degradation, specifically by about 2.6\% in MAE in the Health dataset. The cross-attention module proves vital for text-frequency alignment when its substitution with summation reduces accuracy, particularly in Agriculture, with a drop in MSE of 11.7\%. The real–imaginary variant also shows consistent degradation, increasing MAE by 2.3\% in the Environment dataset, confirming the benefit of modeling spectral interactions jointly of FreqMLPs.

When textual data is excluded, performance deteriorates significantly across all domains, with particularly pronounced effects in the Environment dataset. Figure~\ref{fig:analysis_example} provides a clear case study of this phenomenon, where predictions without text deviate substantially from ground truth in the time domain (Figure~\ref{fig:analysis_example}). In contrast, predictions with text track the ground truth more closely. The frequency-domain comparison in Figure~\ref{fig:analysis_example}b further illustrates this difference, showing how text-enhanced predictions better match the amplitude distribution across frequency bands.  Conversely, removing time series data results in the most dramatic performance drops, most significantly in the Health domain with 30.7\% reduction in MSE.

\textbf{Visualization of Cross-modality Fusion.}
Figure~\ref{fig:analysis_attention_map} illustrates how FreqCMF modulates time-series frequency components in the Climate dataset based on textual context.  The right figure displays frequency amplitude changes in one dimension before and after the FreqCMF block. We observe that the middle-frequency bands (2-7) show more substantial amplitude increases compared to higher frequencies, suggesting that textual information in this domain primarily influences medium-term dynamics. The centre attention heatmap reveals which documents influence specific frequency bands. It shows that documents 13, 16, and 20, containing specific information about ``warm-than-average temperatures'', ``extreme drought conditions'', ``abnormally dry conditions'', and ``moderate-to-exceptional drought'', make a strong impact on the middle-frequency band. In contrast, documents 0 and 5, which do not contain valuable information, contribute less to the frequency components. This demonstrates SpecTF's ability to extract the relevant texts with different frequency bands.


\begin{table}[t]
\renewcommand{\arraystretch}{1.5}  

    \centering
    \caption{Number of trainable parameters, MACs and inference time of models under forecast horizon of 48 in Energy dataset. The best results are shown in \textbf{bold}.}
    \label{tab:model_efficiency}
    \resizebox{0.5\textwidth}{!}{%
    \begin{tabular}{l|c|c|c}
        \cline{1-4}
        \textbf{Models} & \textbf{Parameters} & \textbf{MACs} & \textbf{Inference time}\\ 
        \cline{1-4}
        SpecTF (ours)   & \textbf{397.1K }             & \textbf{1.8M}      & \textbf{9.1ms}    \\ 
        \cline{1-4}
        TimeXer         & 1.72M             & 11.1M        & 88.4ms \\ 
        \cline{1-4}
        MM-TSF          & 6.41M               & 8.23M     & 14.1ms   \\ 
        \cline{1-4}
        TaTS            & 6.41M               & 84.23M  & 41.3ms       \\ 
        \cline{1-4}
    \end{tabular}
    }
\label{tab:model_efficiency}    
\end{table}
\textbf{Model Efficiency.}
Table~\ref{tab:model_efficiency} shows that our SpecTF uses significantly fewer trainable parameters than TimeXer and far fewer than MM-TSF and TaTS. This leads to much lower computational costs, as shown by Multiply-Accumulate Operations (MACs) metric that counts the total number of multiplication and addition operations in a neural network. Besides, SpecTF achieves superior inference speed, outperforming all baselines. These results demonstrate that SpecTF improves efficiency without compromising performance, making it suitable for both resource-limited and large-scale applications.


\section{Conclusion}
\label{section:conclusion}
In conclusion, SpecTF demonstrates that modeling textual influences through frequency-domain decomposition enables more effective modeling of multiscale temporal patterns than traditional time-domain approaches. By fusing text embeddings with spectral components via cross-attention, our framework captures how language data modulates both short-term fluctuations and long-term trends, achieving superior performance across diverse datasets with considerably fewer parameters than leading baselines. SpecTF delivers the greatest benefit when the accompanying text exhibits multiscale temporal influence and offers targeted fusion by reweighting and phase-aligning only those frequency bands modulated by the text.

\section*{Acknowledgment}
This research was funded (partially or fully) by the Australian Government through the Australian Research Council.
Dr Hung Le is the recipient of an Australian Research Council Discovery Early Career Researcher Award (project number DE250100355) funded by the Australian Government.

\bibliographystyle{plainnat}
\bibliography{ref}

\clearpage

\section*{Checklist}

The checklist follows the references. For each question, choose your answer from the three possible options: Yes, No, Not Applicable.  You are encouraged to include a justification to your answer, either by referencing the appropriate section of your paper or providing a brief inline description (1-2 sentences). 
Please do not modify the questions.  Note that the Checklist section does not count towards the page limit. Not including the checklist in the first submission won't result in desk rejection, although in such case we will ask you to upload it during the author response period and include it in camera ready (if accepted).

\textbf{In your paper, please delete this instructions block and only keep the Checklist section heading above along with the questions/answers below.}

\begin{enumerate}

  \item For all models and algorithms presented, check if you include:
  \begin{enumerate}
    \item A clear description of the mathematical setting, assumptions, algorithm, and/or model. [Yes, please refer to Appendix~\ref{sec:theoretical_analysis}]
    \item An analysis of the properties and complexity (time, space, sample size) of any algorithm. [Yes, the analysis of complexity is described in the Section~\ref{section:experiments}]
    \item (Optional) Anonymized source code, with specification of all dependencies, including external libraries. [Yes, we provide source code and guidelines in the supplementary.]
  \end{enumerate}

  \item For any theoretical claim, check if you include:
  \begin{enumerate}
    \item Statements of the full set of assumptions of all theoretical results. [Yes, please refer to Appendix~\ref{sec:theoretical_analysis}]
    \item Complete proofs of all theoretical results. [Yes, please refer to Appendix~\ref{sec:theoretical_analysis}]
    \item Clear explanations of any assumptions. [Yes, please refer to Appendix~\ref{sec:theoretical_analysis}]
  \end{enumerate}

  \item For all figures and tables that present empirical results, check if you include:
  \begin{enumerate}
    \item The code, data, and instructions needed to reproduce the main experimental results (either in the supplemental material or as a URL). [Yes, we provide source code, data and guidelines in the supplementary.]
    \item All the training details (e.g., data splits, hyperparameters, how they were chosen). [Yes, please refer to Appendix~\ref{section:appendix-experiment-details}]
    \item A clear definition of the specific measure or statistics and error bars (e.g., with respect to the random seed after running experiments multiple times). [Yes, please refer to Appendix~\ref{section:appendix-experiment-results}]
    \item A description of the computing infrastructure used. (e.g., type of GPUs, internal cluster, or cloud provider). [Yes, we provide details the infrastructure used in Appendix~\ref{section:appendix-experiment-details}]
  \end{enumerate}

  \item If you are using existing assets (e.g., code, data, models) or curating/releasing new assets, check if you include:
  \begin{enumerate}
    \item Citations of the creator If your work uses existing assets. [Yes]
    \item The license information of the assets, if applicable. [Yes]
    \item New assets either in the supplemental material or as a URL, if applicable. [Not Applicable, we do not introduce any new assets]
    \item Information about consent from data providers/curators. [Yes]
    \item Discussion of sensible content if applicable, e.g., personally identifiable information or offensive content. [No]
  \end{enumerate}

  \item If you used crowdsourcing or conducted research with human subjects, check if you include:
  \begin{enumerate}
    \item The full text of instructions given to participants and screenshots. [Not Applicable]
    \item Descriptions of potential participant risks, with links to Institutional Review Board (IRB) approvals if applicable. [Not Applicable]
    \item The estimated hourly wage paid to participants and the total amount spent on participant compensation. [Not Applicable]
  \end{enumerate}

\end{enumerate}

\clearpage
\appendix
\thispagestyle{empty}

\onecolumn
\aistatstitle{Appendix for \\
    \textit{Spectral Text Fusion: A Frequency-Aware Approach to Multimodal Time-Series Forecasting}}

\section{Theoretical Analysis}
\label{sec:theoretical_analysis}

This section provides formal theoretical justification for why frequency-domain fusion offers principled advantages over time-domain approaches in multimodal time series forecasting.

\subsection{Complex Multiplication Properties}

When two complex numbers are multiplied, the operation simultaneously affects both amplitude and phase characteristics. Given two complex numbers $X_1 = A_1e^{j\phi_1}$ and $X_2 = A_2e^{j\phi_2}$, their multiplication yields:

\begin{equation}
X = X_1 \cdot X_2 = |A_1A_2|e^{j(\phi_1+\phi_2)}
\label{eq:complex_mult}
\end{equation}

where $|A_1A_2|$ represents the combined amplitude modulation and $(\phi_1+\phi_2)$ represents the combined phase shift.

In our SpecTF framework, this property enables \textbf{dual-modal spectral modulation} because textual information can simultaneously influence both the energy distribution across frequencies (amplitude characteristics) and the temporal alignment of frequency components (phase characteristics) of time series data. This is mathematically impossible to achieve with equivalent efficiency in the time domain, where amplitude and phase modifications require separate operations.

\subsection{Convolution Theorem and Global Influence}

The theoretical superiority of frequency-domain fusion is further established by the \textbf{convolution theorem}, which states that convolution in the time domain equals multiplication in the frequency domain:

\begin{equation}
\mathcal{F}\{x(t) * h(t)\} = X(\omega) \cdot H(\omega)
\label{eq:convolution_theorem}
\end{equation}

where $\mathcal{F}$ denotes the Fourier transform, $*$ represents convolution, and $\cdot$ represents point-wise multiplication.

This theorem reveals why our frequency-domain text embedding can modulate every timestep of the time series in one unified operation, rather than relying on local, step-wise interactions typical in time-domain approaches. The mathematical equivalence ensures that our multiplication fusion $\hat{X}_{fuse} = \hat{X}_{emb} \odot \hat{O}$ corresponds to a global convolution operation across all temporal positions simultaneously.

\subsection{Energy Conservation Guarantees}

\textbf{Parseval's Energy Conservation Theorem} provides mathematical guarantees for information preservation in our approach:

\begin{equation}
\int_{-\infty}^{\infty} |x(t)|^2 dt = \frac{1}{2\pi} \int_{-\infty}^{\infty} |X(f)|^2 df
\label{eq:parseval}
\end{equation}

This theorem ensures that our frequency-domain operations preserve signal energy, providing a mathematical guarantee that no information is lost during domain transformation. The energy conservation property is crucial for maintaining the integrity of both temporal patterns and textual influences throughout the fusion process.

\subsection{Multi-Scale Influence Modeling}

Time-domain fusion approaches face fundamental limitations in modeling multi-scale temporal influences. Consider a textual signal $s(t)$ that needs to influence a time series $x(t)$ at multiple temporal scales. In the time domain, this requires:

\begin{equation}
y(t) = \sum_{k} \alpha_k \cdot s(t) * h_k(t) * x(t)
\label{eq:time_domain_multiscale}
\end{equation}

where $h_k(t)$ represents different temporal scale filters and $\alpha_k$ are scale-specific weights. This formulation requires multiple convolution operations and explicit scale decomposition.

In contrast, our frequency-domain approach naturally decomposes signals into scale-specific components through the Fourier transform, where each frequency bin $k$ corresponds to a specific temporal scale $\frac{2\pi}{k}$. The fusion operation:

\begin{equation}
\hat{Y}(k) = \hat{X}(k) \cdot \hat{S}(k)
\label{eq:freq_domain_fusion}
\end{equation}

simultaneously addresses all temporal scales with a single multiplication operation per frequency component.

\section{Implementation Details}
\label{section:appendix-experiment-details}
\subsection{Dataset}

Our experimental evaluation utilizes two comprehensive multimodal time series benchmarks with distinct characteristics and evaluation protocols.

\textbf{Time-MMD Benchmark.} We employ the Time-MMD dataset comprising 9 multimodal datasets across daily, weekly, and monthly frequencies, with standardized temporal parameters adapted from established forecasting protocols. Following the setup in \citep{DBLP:journals/corr/abs-2502-08942}, we adopt a fixed 24-step lookback window across all domains to enable consistent cross-domain comparisons. Prediction horizons follow frequency-specific configurations: daily datasets use horizons {48, 96, 192, 336} steps, weekly datasets employ {12, 24, 36, 48} steps, and monthly datasets utilize {6, 8, 10, 12} steps. Table~\ref{tab:dataset} describes the statistics of each and the corresponding lookback window and forecasting horizon.

\textbf{TimeText Corpus (TTC) Benchmark.} For the TTC evaluation, we conduct experiments on the Medical and Climate datasets following the standardized protocol established in \citep{kim2024multimodalforecasterjointlypredicting}. We maintain consistency by adopting a 24-step lookback window for both datasets, with prediction horizons configured as {12, 24, 36, 48} steps across all scenarios. 

\subsection{Code and Reproducibility}
The experimental implementation leverages NVIDIA V100 GPUs for training, with all models undergoing three independent runs using random seeds to ensure statistical reliability through averaged metrics. Hyperparameter configurations follow the specifications detailed in Table~\ref{tab:hyperparam}. The codebase, implemented in PyTorch 1.13 with CUDA 11.7 dependencies, along with comprehensive installation instructions and environment setup guidelines, is provided in the supplementary materials to ensure reproducibility. 

\subsection{Metrics}
Our implementation utilizes Mean Squared Error (MSE) as the primary optimization objective and Mean Absolute Error (MAE) as the key evaluation metric, aligning with established practices for regression tasks in machine learning literature

The MSE loss function calculates the average squared difference between predicted values ($\hat{y}_i$) and ground truth labels ($y_i$) across N samples:
\begin{equation}
    \centering
    \text{MSE} = \frac{1}{N}\sum_{i=1}^N(y_i-\hat{y}_i)^2
\end{equation}

This quadratic penalty strongly penalizes large prediction errors, making it particularly effective for gradient-based optimization landscapes. For final model evaluation, MAE provides an interpretable measure of average absolute deviation:

\begin{equation}
    \centering
    \text{MAE} = \frac{1}{N}\sum_{i=1}^N|y_i-\hat{y}_i|^2
\end{equation}
This MAE metric's linear scaling makes it robust to outliers compared to MSE, providing complementary insights into model performance. 

Specifically, in our problem, MSE and MAE are computed as follows:
\begin{equation}
    \centering
    \begin{aligned}
        \text{MSE} = \frac{1}{H}\sum_{i=1}^H(y_i-x^p_i)^2 \\
        \text{MAE} = \frac{1}{H}\sum_{i=1}^H|y_i-x^p_i|^2
    \end{aligned}
\end{equation}
where $Y=[y_1,\dots\,y_H]$ is the ground truth, $X^p=[x^p_1,\dots\,x^p_H]$ is the prediction, $H$ is the forecast horizon.
\begin{table}[t]
    \centering
    \caption{Default parameters of SpecTF.}
    \resizebox{0.8\textwidth}{!}{
    \begin{tabular}{|l|l|c|}
    \toprule
     \textbf{Hyperparameter} & \textbf{Description} & \textbf{Choices}\\
     \midrule
     batch\_size & The batch size for training & 32 \\
     seq\_len & Lookback window length & 24 \\
     prior\_weight & Weight for combination 
 & \{0.1 0.2 0.4 0.5\} \\
     train\_epochs & Number of training epochs & 50 \\
     patience & Early stopping patience & 20 \\
     pool\_type & Pooling type for text embedding & avg \\
     dropout & Dropout & 0.1 \\
    \midrule

    \end{tabular}
    }
    \label{tab:hyperparam}
\end{table}

\begin{table}[t]
    \centering
    \caption{Overview of numerical data in Time-MMD and the lookback window, forecasting horizon in our implementation.}
    \resizebox{0.8\textwidth}{!}{
    \begin{tabular}{|c|c|c|c|c|}
    \toprule
     \textbf{Dataset} & \textbf{Frequency} & \textbf{Number of samples} & \textbf{Lookback window} & \textbf{Forecasting horizon} \\
     \midrule
     Agriculture & Monthly & 496 & 24 &  \{6, 8, 10, 12\} \\
    \midrule
     Climate & Monthly & 496 & 24 &  \{6, 8, 10, 12\} \\
     \midrule
     Economy & Monthly & 423 & 24 &  \{6, 8, 10, 12\} \\
     \midrule
     Energy & Weekly & 1479 & 24 &   \{12, 24, 36, 48\}\\
     \midrule
     Environment & Daily & 11102 & 24 &   \{48, 96, 192, 336\}\\
     \midrule
     Health & Weekly & 1389 & 24 &    \{12, 24, 36, 48\}\\
     \midrule
     Security & Monthly & 297 & 24 &  \{6, 8, 10, 12\} \\
     \midrule
     Social Good & Monthly & 900 & 24 &  \{6, 8, 10, 12\} \\
     \midrule
     Traffic & Monthly & 531 & 24 &  \{6, 8, 10, 12\} \\
    \bottomrule

    \end{tabular}
    }
    \label{tab:dataset}
\end{table}

\section{Full Experiment Results}
\label{section:appendix-experiment-results}
\subsection{Forecasting Results on Time-MMD benchmark}
\label{section:appendix-forecasting-results}
We provide full forecasting results in Table~\ref{tab:full_forecasting_results} at multiple forecast horizons as described in Appendix~\ref{section:appendix-experiment-details}. Overall, SpecTF outperforms both unimodal and multimodal baselines across 8 in 9 datasets at different prediction length.
\clearpage
\begin{table}[t]
\vspace{-5pt}
  \caption{Full forecasting results across nine datasets in TimeMMD benchmark \citep{liu2024timemmd} at 
multiple forecast horizons.}\label{tab:full_forecasting_results}
  \vskip 0.05in
  \centering
  \begin{small}
  \renewcommand{\multirowsetup}{\centering}
  \setlength{\tabcolsep}{4.1pt}
  \resizebox{0.95\textwidth}{!}{
  \begin{tabular}{cc|cc|cc|cc|cc|cc|cc|cc}
    \toprule
    \multicolumn{2}{c|}{\multirow{2}{*}{{Models}}} & 
    \multicolumn{2}{c|}{SpecTF (ours)} &
    \multicolumn{2}{c|}{TaTS} &
    \multicolumn{2}{c|}{MM-TSF} &  \multicolumn{2}{c|}{TimeXer} & \multicolumn{2}{c|}{FreTS} & \multicolumn{2}{c|}{TimeLLM} & \multicolumn{2}{c}{ChatTime}\\
    \cmidrule(lr){3-16}
    \multicolumn{2}{c|}{Method}  & \scalebox{1.0}{MSE} & \scalebox{1.0}{MAE}  & \scalebox{1.0}{MSE} & \scalebox{1.0}{MAE}  & \scalebox{1.0}{MSE} & \scalebox{1.0}{MAE} & \scalebox{1.0}{MSE} & \scalebox{1.0}{MAE} & \scalebox{1.0}{MSE} & \scalebox{1.0}{MAE}  & \scalebox{1.0}{MSE} & \scalebox{1.0}{MAE}  & \scalebox{1.0}{MSE} & \scalebox{1.0}{MAE}  \\
    \toprule
    \multirow{5}{*}{\scalebox{1.0}{Agriculture}}
    & 6 & 0.064 & 0.170 & 0.075 & 0.196 & 0.084 & 0.216 & 0.088 & 0.211 & 0.082 & 0.201 & 0.092 & 0.218 & 0.193 & 0.283 \\
    & 8 & 0.084 & 0.192 & 0.097 & 0.219 & 0.112 & 0.253 & 0.109 & 0.239 & 0.101 & 0.229  & 0.151 & 0.303 & 0.264 & 0.320 \\
    & 10 & 0.110 & 0.236 & 0.127 & 0.254 & 0.133 & 0.276 & 0.144 & 0.2811 & 0.125 & 0.262 & 0.176 & 0.330 & 0.345 & 0.371\\
    & 12 & 0.152 & 0.275 & 0.167 & 0.291 & 0.165 & 0.305 & 0.187 & 0.322 & 0.172 & 0.303 & 0.210 & 0.354 & 0.418 &  0.407 \\
    \cmidrule(lr){2-16}
    & Avg  & 0.103  & 0.218 & 0.112 & 0.237 & 0.123 & 0.263 & 0.132 & 0.263 & 0.120 & 0.249 & 0.157 & 0.301 & 0.305 & 0.345  \\    
    \midrule
    \multirow{5}{*}{\scalebox{1.0}{Climate}}
    & 6 & 0.904 & 0.748 & 0.990 & 0.780 & 1.074 & 0.822 & 1.027 & 0.794 & 0.286 & 0.878 & 1.126 & 0.854 & 1.490 & 0.981  \\
    & 8 & 0.927 & 0.756 & 0.998 & 0.786 & 1.067 & 0.823 & 1.059 & 0.804 & 0.283 & 0.883 & 1.165 & 0.866 & 1.530 & 1.006  \\
    & 10 & 0.965 & 0.775 & 1.010 & 0.794 & 1.100 & 0.839  & 1.067 & 0.812 & 0.294 & 0.893 & 1.200 & 0.877 & 1.519 & 1.005 \\
    & 12 & 0.960 & 0.770 & 1.012 & 0.798 & 1.101 & 0.842 & 1.058 & 0.810 & 0.293 & 0.897 & 1.210 & 0.875 & 1.570 & 1.019  \\
    \cmidrule(lr){2-16}
    & Avg  & 0.938  & 0.762 & 1.002 & 0.790 & 1.053 & 0.805 & 1.053 & 0.805 & 1.289 & 0.888 & 1.175 & 0868 & 1.528 & 1.002 \\    
    \midrule
    \multirow{5}{*}{\scalebox{1.0}{Economy}}
    & 6 & 0.0084 & 0.0770 & 0.0080 & 0.0759 & 0.0116 & 0.0856 & 0.0133 & 0.0935 & 0.0150 & 0.0975 & 0.0307 & 0.1414 & 0.478 & 0.169  \\
    & 8 & 0.0086 & 0.0785 & 0.0083 & 0.0778 & 0.0117 & 0.0880 & 0.0140 & 0.0958 & 0.0149 & 0.0981 & 0.0310 & 0.1435 & 0.056 & 0.179  \\
    & 10 & 0.0085 & 0.0779 & 0.0083 & 0.0773 & 0.0126 & 0.0906 & 0.0143 & 0.0952 & 0.0155 & 0.0992 & 0.0323 & 0.1444 & 0.056 & 0.181  \\
    & 12 & 0.0086 & 0.0784 & 0.0084 & 0.0767 & 0.0125 & 0.0891 & 0.0148 & 0.0982 & 0.0154 & 0.0996 & 0.0336 & 0.1460 & 0.059 & 0.184 \\
    \cmidrule(lr){2-16}
    & Avg  & 0.0085  & 0.0780 & 0.0083 & 0.0770 & 0.0096 & 0.0817 & 0.0096 & 0.0817 & 0.0152 & 0.0986  & 0.0319 & 0.1438 & 0.0547 & 0.1786 \\    
    \midrule
    \multirow{5}{*}{\scalebox{1.0}{Energy}}
    & 12 & 0.096 & 0.219 & 0.103 & 0.227 & 0.117 & 0.256 &  0.137 & 0.278  & 0.127 & 0.251  & 0.105 & 0.225 & 0.135 & 0.282 \\
    & 24 & 0.199 & 0.326 & 0.206 & 0.336 & 0.210 & 0.355 &  0.242 & 0.369  & 0.245 & 0.373  & 0.216 & 0.340 & 0.238 & 0.369 \\
    & 36 & 0.282 & 0.396 & 0.313 & 0.420 & 0.307 & 0.431 &  0.324 & 0.429  & 0.337 & 0.433 & 0.307 & 0.408 & 0.328 & 0.421 \\
    & 48 & 0.406 & 0.493 & 0.433 & 0.514 & 0.447 & 0.520 & 0.447 & 0.523 & 0.487 & 0.548  & 0.427 & 0.503 & 0.435 & 0.516 \\
    \cmidrule(lr){2-16}
    & Avg  & 0.246  & 0.359 & 0.264 & 0.374 & 0.270 & 0.393 & 0.288 & 0.400 & 0.299 & 0.401  & 0.264 & 0.369 & 0.284 & 0.397 \\    
    \midrule
    \multirow{5}{*}{\scalebox{1.0}{Environment}}
    & 48 & 0.265 & 0.370 & 0.267 & 0.374 & 0.288 & 0.391 & 0.270 & 0.374 & 0.350 & 0.464  & 0.314 & 0.393 & 0.342  & 0.429 \\
    & 96 & 0.265 & 0.368 & 0.268 & 0.371 & 0.294 & 0.392 & 0.280 & 0.376 & 0.367 & 0.479  & 0.352 & 0.414 & 0.351 & 0.428 \\
    & 192 & 0.255 & 0.365 & 0.268 & 0.367 & 0.292 & 0.398 & 0.271 & 0.367 & 0.385 & 0.492  & 0.354 & 0.410 & 0.345 & 0.423 \\
    & 336 & 0.253 & 0.366 & 0.262 & 0.368 & 0.291 & 0.397 & 0.267 & 0.367 & 0.386 & 0.493  & 0.348 & 0.426 & 0.342 & 0.424 \\
    \cmidrule(lr){2-16}
    & Avg  & 0.259  & 0.367 & 0.266 & 0.370 & 0.391 & 0.494 & 0.272 & 0.371 & 0.372 & 0.482  & 0.342 & 0.411 & 0.345 & 0.426 \\    
    \midrule
    \multirow{5}{*}{\scalebox{1.0}{Health}}
    & 12 & 0.985 & 0.647 & 0.979 & 0.663 & 1.196 & 0.828 & 1.181 & 0.744 & 1.141 & 0.680  & 1.308 & 0.752 & 1.382 & 0.773 \\
    & 24 & 1.256 & 0.718 & 1.297 & 0.741 & 1.440 & 0.865 & 1.402 & 0.786 & 1.546 & 0.785  & 1.724 & 0.868 & 1.758 & 0.888 \\
    & 36 & 1.395 & 0.769 & 1.442 & 0.787 & 1.669 & 0.920 & 1.455 & 0.803 & 1.691 & 0.881 & 1.869 & 0.914 & 1.904 & 0.938 \\
    & 48 & 1.470 & 0.797 & 1.519 & 0.820 & 1.755 & 0.973 & 1.538 & 0.841 & 1.790 & 0.910  & 1.895 & 0.949 & 1.930 & 0.973 \\
    \cmidrule(lr){2-16}
    & Avg  & 1.276  & 0.733 & 1.340 & 0.763 & 1.508 & 0.897 & 1.394 & 0.973 & 1.542 & 0.814  & 1.699 & 0.871 & 1.730 & 0.893 \\    
    \midrule
    \multirow{5}{*}{\scalebox{1.0}{Security}}
    & 6 & 106.364 & 4.507 & 108.460 & 4.929 & 113.759 & 5.418 & 106.859 & 4.754 & 126.132 & 6.265 & 106.535 & 4.694 & 113.701  & 5.835 \\
    & 8 & 106.947 & 4.769 & 110.180 & 5.042 & 115.563 & 5.542 & 109.134 & 4.855 & 129.167 & 6.215  & 108.205 & 4.718 & 117.083 & 5.877 \\
    & 10 & 109.767 & 4.616 & 111.352 & 5.104 & 116.793 & 5.611 & 110.484 & 4.915 & 129.848 & 6.237  & 109.566 & 4.905 & 123.805 & 6.009 \\
    & 12 & 110.567 & 5.089 & 112.405 & 5.201 & 117.897 & 5.717 & 111.411 & 4.956 & 130.346 & 6.322  & 110.865 & 5.028 & 190.503 & 6.219 \\
    \cmidrule(lr){2-16}
    & Avg  & 108.411  & 4.745 & 110.600 & 5.069 & 116.003 & 5.572 & 109.472 & 4.870 & 128.873 & 6.260  & 108.812 & 4.836 &  136.273 &  5.985 \\    
    \midrule
    \multirow{5}{*}{\scalebox{1.0}{Social Good}}
    & 6 & 0.907 & 0.404 & 0.943 & 0.388 & 1.088 & 0.469 & 0.924 & 0.437 & 1.008 & 0.469  & 0.888 & 0.432 & 1.193 & 0.487 \\
    & 8 & 0.939 & 0.421 & 1.012 & 0.426 & 1.183 & 0.499 & 0.978 & 0.459 & 1.017 & 0.500 & 0.916 & 0.473 & 1.117 & 0.493 \\
    & 10 & 0.962 & 0.434 & 0.996 & 0.448 & 1.198 & 0.505 & 1.018 & 0.485 & 1.145 & 0.541  & 1.039 & 0.527 & 1.241 & 0.511\\
    & 12 & 1.040 & 0.514 & 1.058 & 0.473 & 1.264 & 0.516 & 1.089 & 0.532 & 1.174 & 0.571  & 1.492 & 0.859 & 1.189 & 0.513\\
    \cmidrule(lr){2-16}
    & Avg  & 0.962  & 0.443 & 1.012 & 0.449 & 1.183 & 0.497 & 1.002 & 0.478 & 1.086 & 0.520 & 1.084 & 0.573 & 1.185 & 0.501 \\    
    \midrule
    \multirow{5}{*}{\scalebox{1.0}{Traffic}}
    & 6 & 0.169 & 0.202 & 0.182 & 0.208 & 0.186 & 0.225 & 0.166 & 0.205 & 0.200 & 0.232 & 0.271 & 0.383 & 0.255 & 0.344 \\
    & 8 & 0.169 & 0.210 & 0.184 & 0.219 & 0.190 & 0.231 & 0.167 & 0.211 & 0.199 & 0.238  & 0.276 & 0.385 & 0.257 & 0.347 \\
    & 10 & 0.173 & 0.218 & 0.190 & 0.230 & 0.195 & 0.241 & 0.176 & 0.219 & 0.205 & 0.241  & 0.279 & 0.385 & 0.259 & 0.347 \\
    & 12 & 0.175 & 0.210 & 0.213 & 0.214 & 0.217 & 0.236 & 0.197 & 0.205 & 0.215 & 0.253 & 0.333 & 0.399 &  0.261 & 0.349 \\
    \cmidrule(lr){2-16}
    & Avg  & 0.171  & 0.210 & 0.192 & 0.218 & 0.197 & 0.233 & 0.176 & 0.210 & 0.205 & 0.241 & 0.289 & 0.388 & 0.258 & 0.347 \\    
    \midrule
  \end{tabular}}
    \end{small}
  \vspace{-5pt}
\end{table}

\subsection{Forecasting Results on TTC benchmark}
\label{section:appendix-forecasting-results_ttc}
We have conducted additional experiments (Table ~\ref{tab:full_forecasting_results_ttc}) on two diverse datasets, Medical and Climate, from the TimeText Corpus (TTC) \citep{kim2024multimodalforecasterjointlypredicting}. These datasets originate from MIMIC-III clinical notes and National Weather Service forecasts, respectively, and introduce different characteristics such as higher noise levels (in medical text) and finer temporal granularity (in weather data).

We follow the same experimental setup with a lookback window of 24 and compare our method against TaTS, the strongest baseline in our main experiments. These results confirm that our frequency-domain fusion generalises well to very different data regimes, reinforcing that the gains reported on Time-MMD are not benchmark-specific.

\begin{table}[t]
\vspace{-5pt}
  \caption{Full forecasting results across nine datasets in TTC benchmark at 
multiple forecast horizons.}\label{tab:full_forecasting_results_ttc}
  \vskip 0.05in
  \centering
  \begin{small}
  \renewcommand{\multirowsetup}{\centering}
  \setlength{\tabcolsep}{4.1pt}
  \resizebox{0.6\textwidth}{!}{
  \begin{tabular}{cc|cc|cc|cc|cc|cc|cc|cc}
    \toprule
    \multicolumn{2}{c|}{\multirow{2}{*}{{Models}}} &
    \multicolumn{2}{c|}{SpecTF (ours)} &
    \multicolumn{2}{c|}{TaTS} &
    \multicolumn{2}{c|}{MM-TSF} & \multicolumn{2}{c|}{TimeXer} & \multicolumn{2}{c|}{FreTS} & \multicolumn{2}{c|}{TimeLLM} & \multicolumn{2}{c}{ChatTime} \\
    \cmidrule(lr){3-16}
    \multicolumn{2}{c|}{Method}  & \scalebox{1.0}{MSE} & \scalebox{1.0}{MAE}  & \scalebox{1.0}{MSE} & \scalebox{1.0}{MAE}  & \scalebox{1.0}{MSE} & \scalebox{1.0}{MAE} & \scalebox{1.0}{MSE} & \scalebox{1.0}{MAE} & \scalebox{1.0}{MSE} & \scalebox{1.0}{MAE} & \scalebox{1.0}{MSE} & \scalebox{1.0}{MAE} & \scalebox{1.0}{MSE} & \scalebox{1.0}{MAE} \\
    \toprule
    \multirow{5}{*}{\scalebox{1.0}{Climate}}
    & 12 & 0.655 & 0.576 & 0.759 & 0.620 & 0.767 & 0.647 & 0.741 & 0.615 & 0.857 & 0.665 & 0.656 & 0.583 & 0.868 & 0.607\\
    & 24 & 0.690 & 0.603 & 0.785 & 0.641 & 0.802 & 0.658 & 0.759 & 0.627 & 0.865 & 0.679 & 0.698 & 0.605 & 0.906 & 0.629 \\
    & 36 & 0.701 & 0.610 & 0.797 & 0.653 & 0.811 & 0.660 & 0.769 & 0.629 & 0.889 & 0.683 & 0.700 & 0.616 & 0.943 & 0.650\\
    & 48 & 0.727 & 0.631 & 0.819 & 0.668 &  0.831 & 0.684 & 0.771 & 0.653 & 0.905 & 0.701 & 0.743 & 0.639 & 0.987 & 0.671 \\
    \cmidrule(lr){2-16}
    & Avg  & 0.693  & 0.605 & 0.790 & 0.636 & 0.802 & 0.662 & 0.760 &  0.631 & 0.878 & 0.682 & 0.699 & 0.611 & 0.926 & 0.639 \\
    \midrule
    \multirow{5}{*}{\scalebox{1.0}{Medical}}
    & 12 & 0.927 & 0.666 & 1.265 & 0.768 &  1.422 & 1.056 & 1.293 & 0.861 & 1.493 & 1.051 & 0.911 & 0.662 & 1.572 & 1.104\\
    & 24 & 1.271 & 0.821 & 1.297 & 0.998 &  1.427 & 1.064 & 1.309 & 0.873 & 1.507 & 1.067 & 1.275 & 0.822 & 1.597 & 1.108\\
    & 36 & 1.441 & 0.888 & 1.332 & 1.017 & 1.443 & 1.077 & 1.317 & 0.885 & 1.519 & 1.075 & 1.483 & 0.895 & 1.622 & 1.126 \\
    & 48 & 1.554 & 0.936 & 1.357 & 1.032 & 1.451 & 1.092 & 1.325 & 0.901 & 1.525 & 1.099 & 1.637 & 0.964 & 1.649 & 1.131 \\
    \cmidrule(lr){2-16}
    & Avg  & 1.298  & 0.828 & 1.313 & 0.953 & 1.436 & 1.072 & 1.311 & 0.880 & 1.511 & 1.073 & 1.327 & 0.836 & 1.610 & 1.117\\ 
    \midrule
  \end{tabular}}
    \end{small}
  \vspace{-5pt}
\end{table}
\subsection{Hyperparameter Study}
\label{section:appendix-hyperparam}

\begin{figure*}[h]
\centering
\vspace{-3mm}
\subfigure[Agriculture]{
\includegraphics[width=0.31\textwidth]{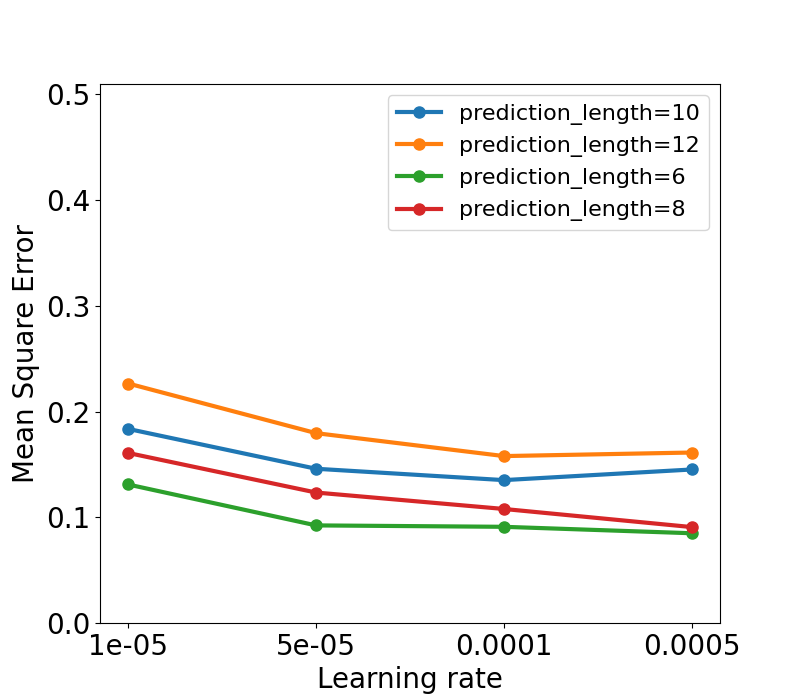}
}
\subfigure[Climate]{
\includegraphics[width=0.31\textwidth]{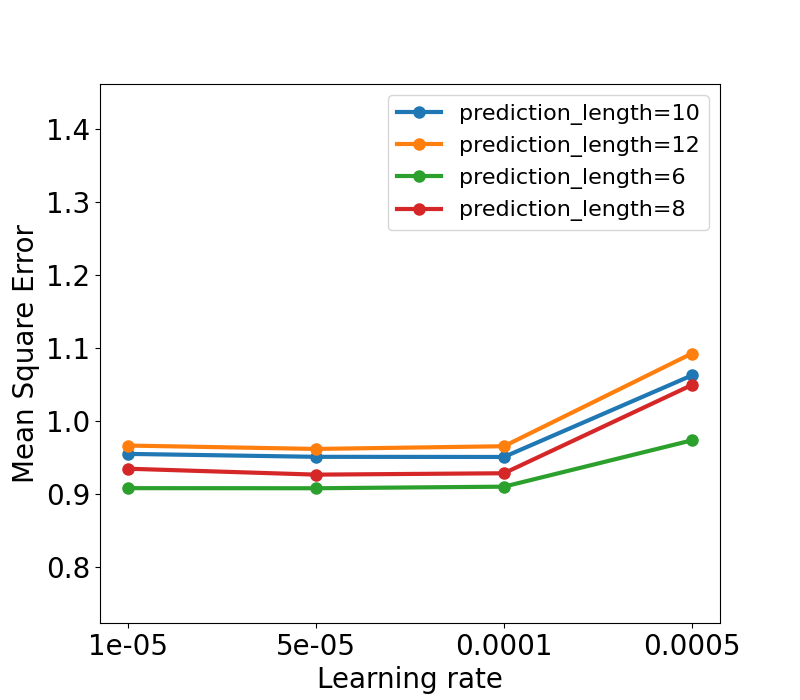}
}
\subfigure[Economy]{
\includegraphics[width=0.31\textwidth]{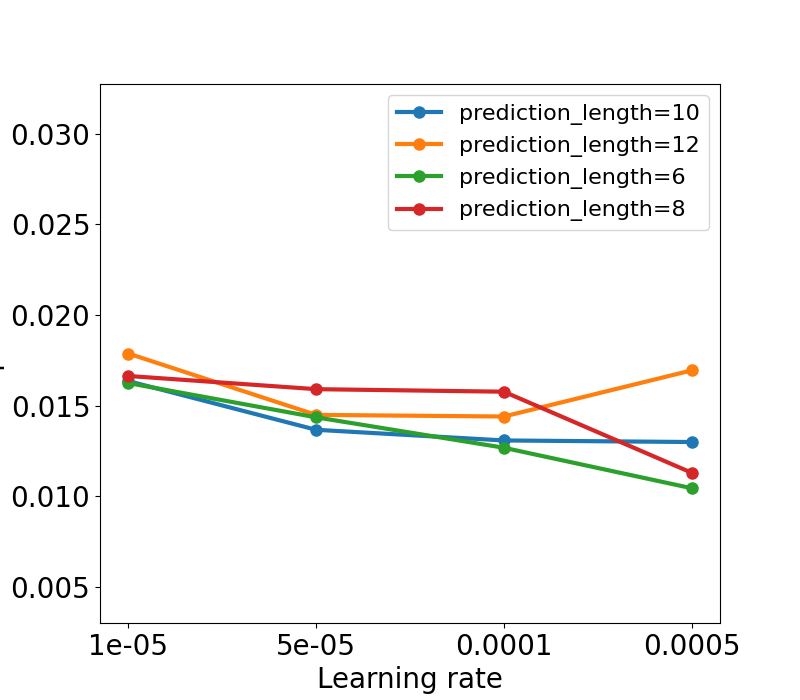}
}
\subfigure[Energy]{
\includegraphics[width=0.31\textwidth]{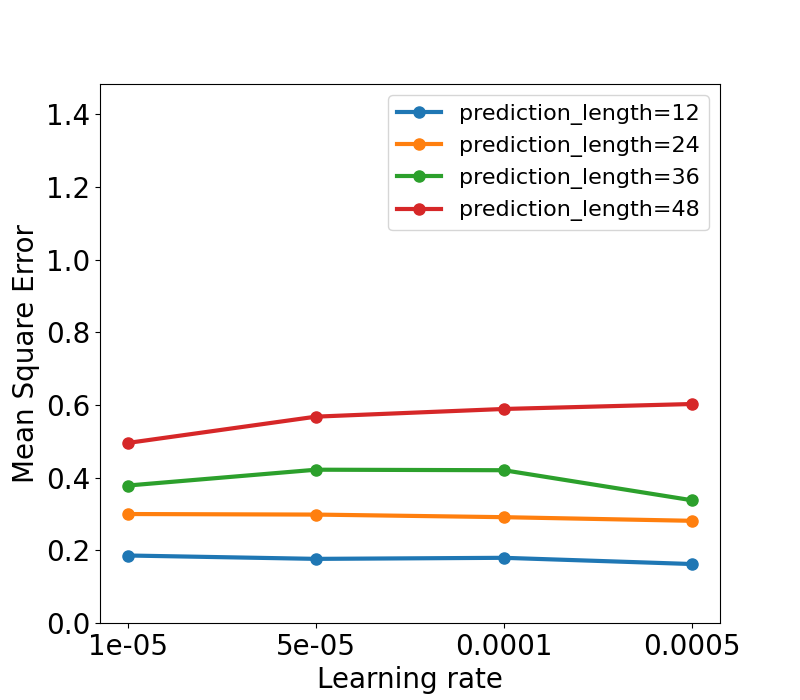}
}
\subfigure[Environment]{
\includegraphics[width=0.31\textwidth]{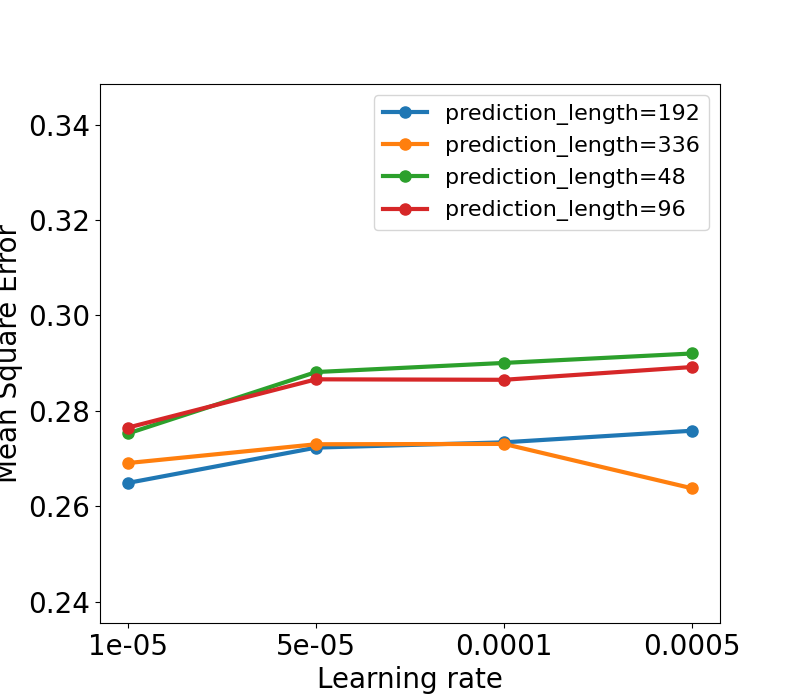}
}
\subfigure[Health]{
\includegraphics[width=0.31\textwidth]{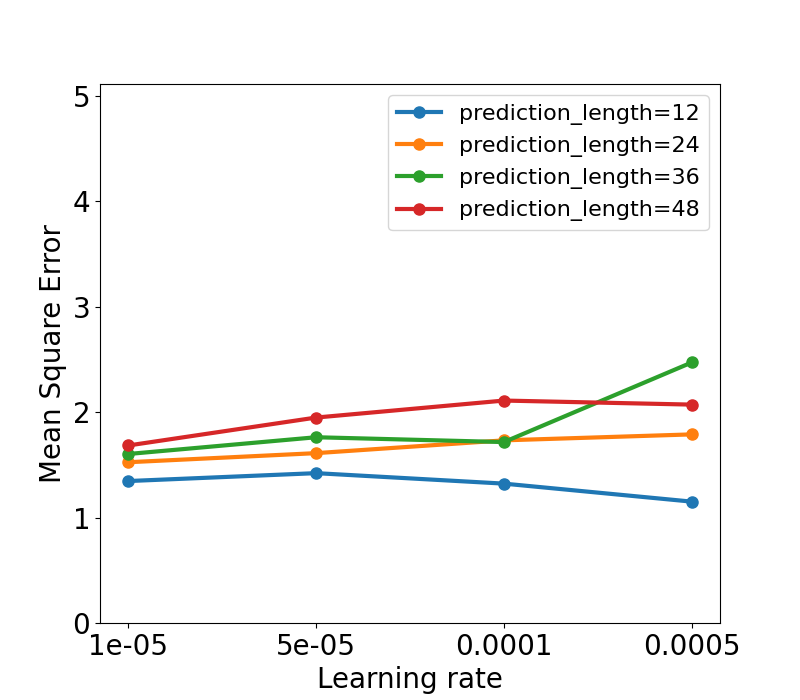}
}
\subfigure[Security]{
\includegraphics[width=0.31\textwidth]{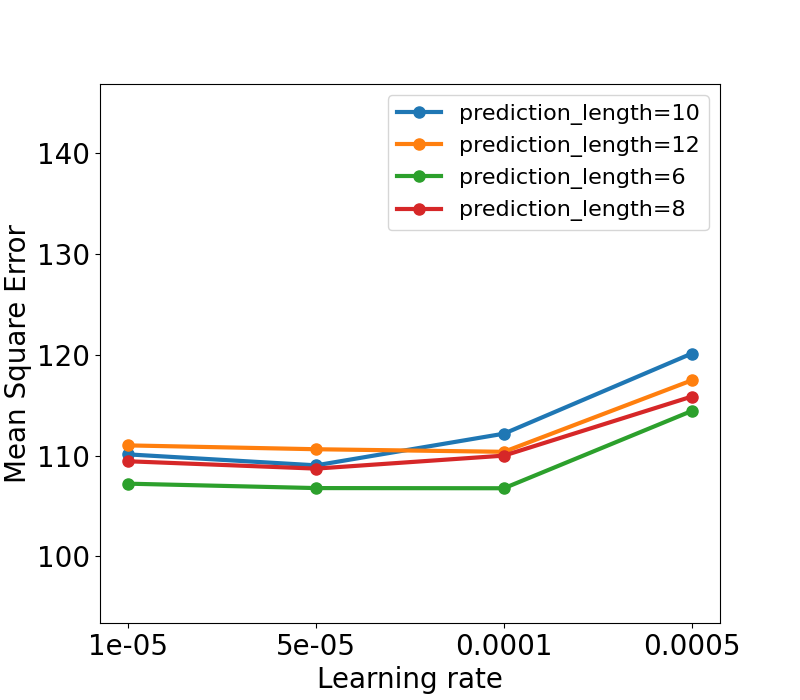}
}
\subfigure[Social Good]{
\includegraphics[width=0.31\textwidth]{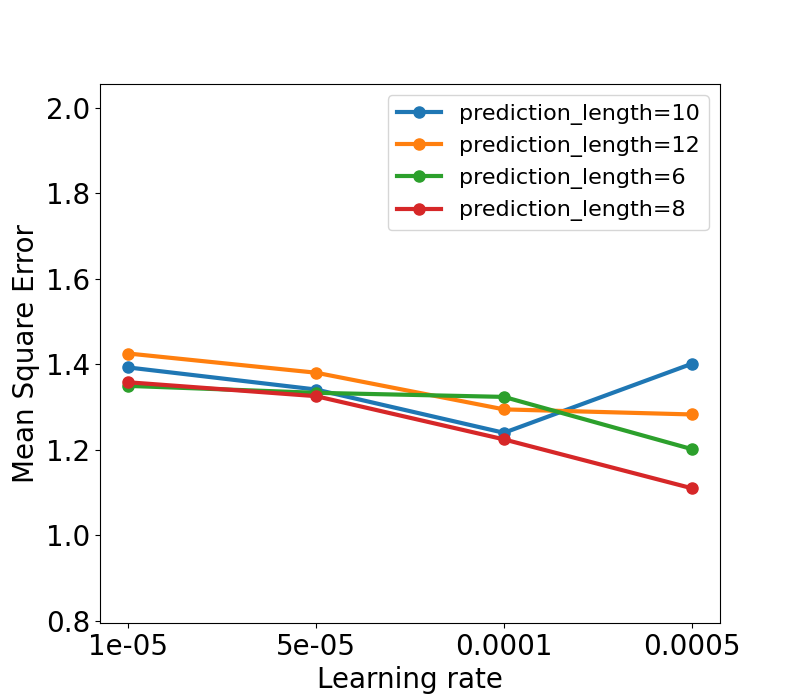}
}
\subfigure[Traffic]{
\includegraphics[width=0.31\textwidth]{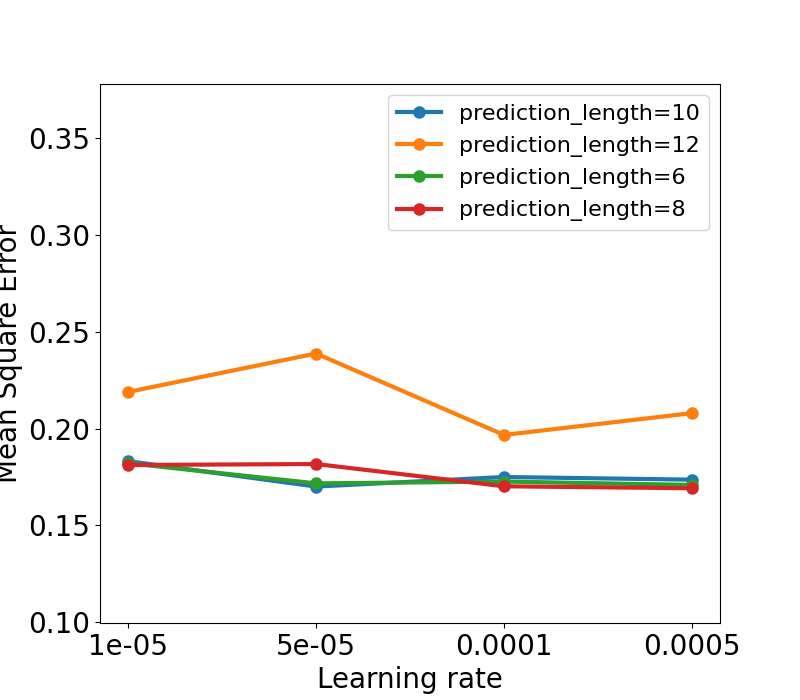}
}
\caption{Parameter study on learning rate.}
\label{fig:param_study_lr}
\vspace{-3mm}
\end{figure*}

Figure~\ref{fig:param_study_lr} provides a comprehensive analysis of the method’s robustness to learning rate choices. Performances remain stable within a broad learning rate range (1e-5 to 5e-4). However, the Security domain shows significant sensitivity, where high rates ($>0.0001$) cause MSE spikes ($>140$), necessitating precise tuning. Long-horizon tasks (e.g., Health 48-step forecasts) also require lower rates to maintain stability, highlighting domain-specific optimization needs despite general adaptability. These edge cases highlight the framework’s default stability while underscoring domain-specific tuning needs for volatile or long-horizon tasks.

\begin{figure*}[h]
\centering
\vspace{-3mm}
\subfigure[Agriculture]{
\includegraphics[width=0.31\textwidth]{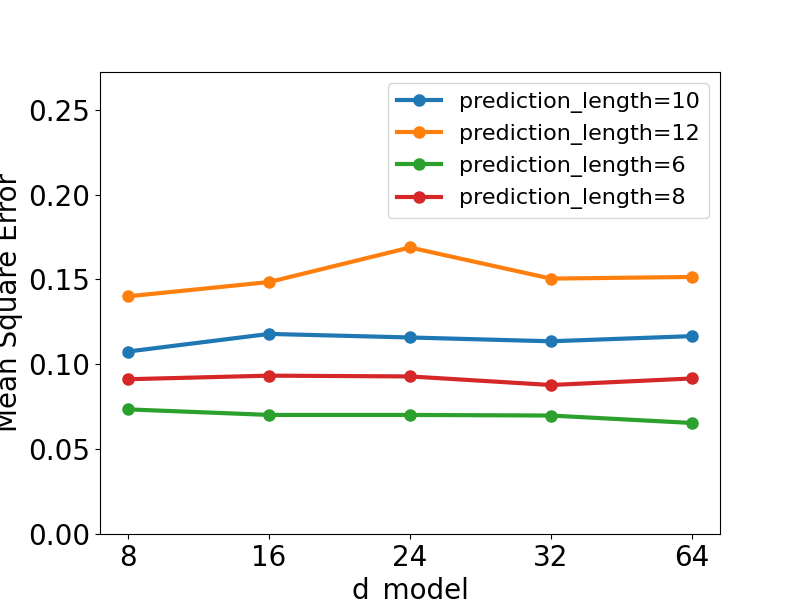}
}
\subfigure[Climate]{
\includegraphics[width=0.31\textwidth]{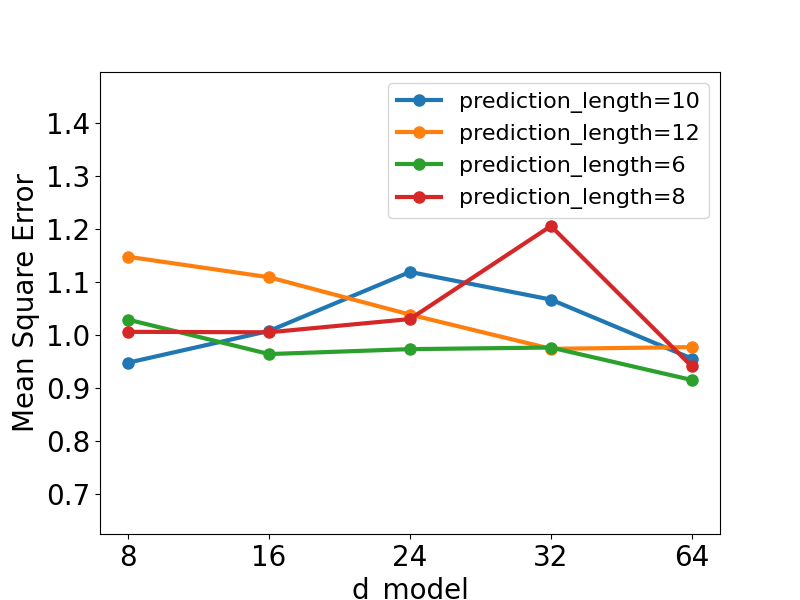}
}
\subfigure[Economy]{
\includegraphics[width=0.31\textwidth]{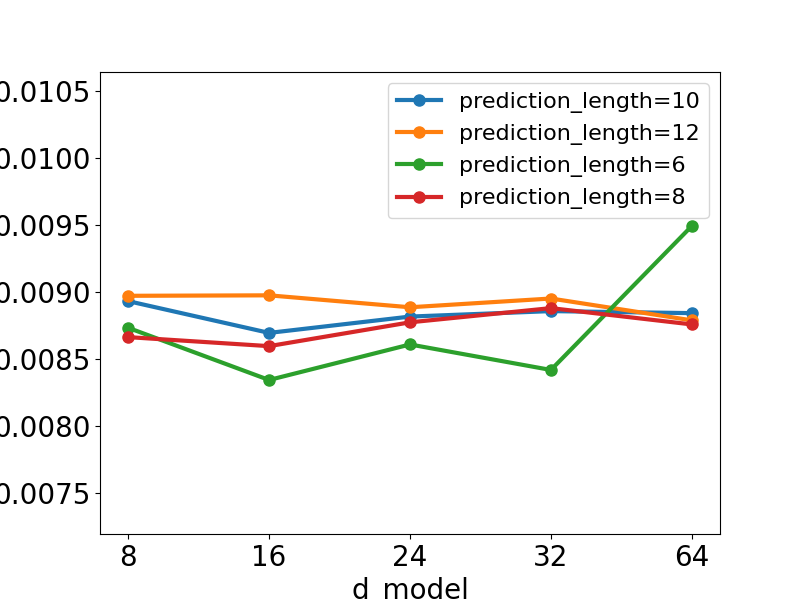}
}
\subfigure[Energy]{
\includegraphics[width=0.31\textwidth]{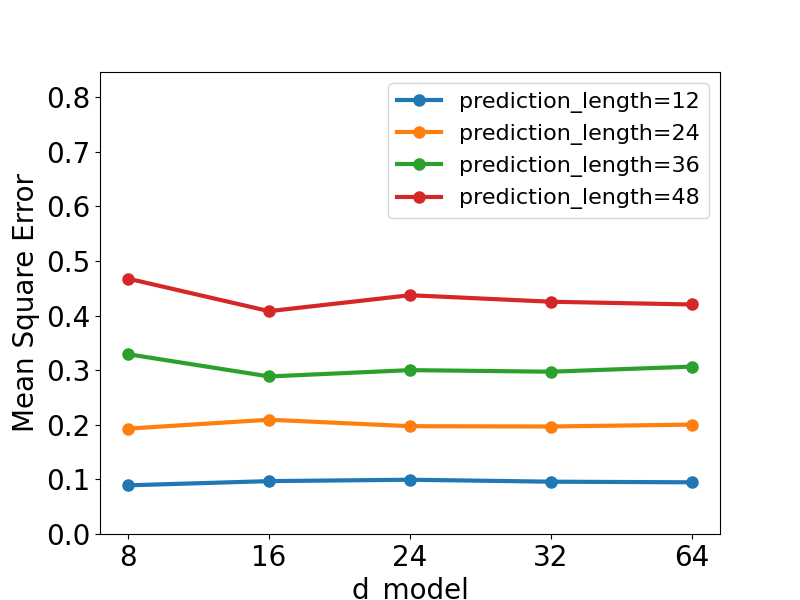}
}
\subfigure[Environment]{
\includegraphics[width=0.31\textwidth]{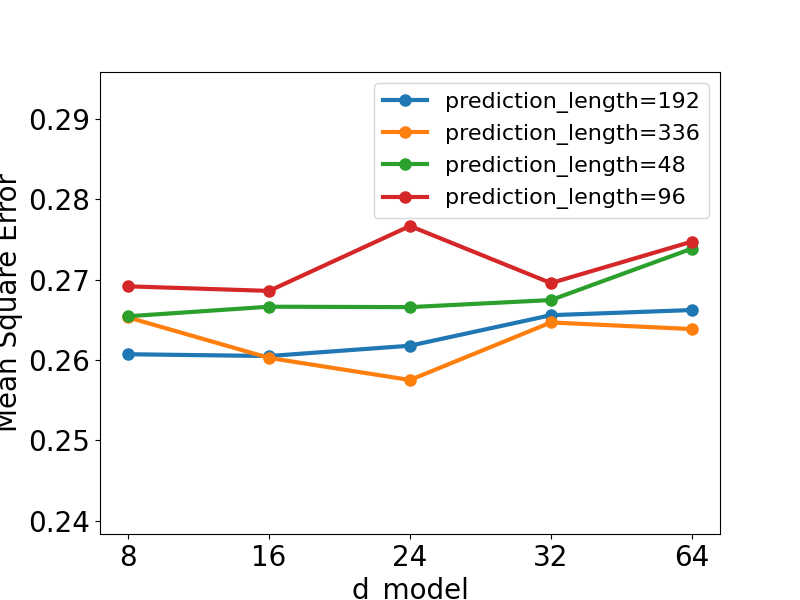}
}
\subfigure[Health]{
\includegraphics[width=0.31\textwidth]{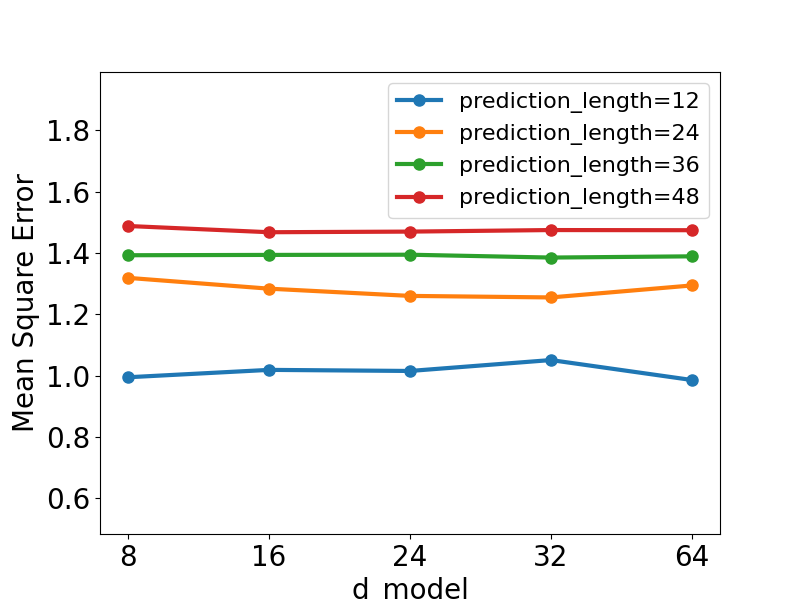}
}
\subfigure[Security]{
\includegraphics[width=0.31\textwidth]{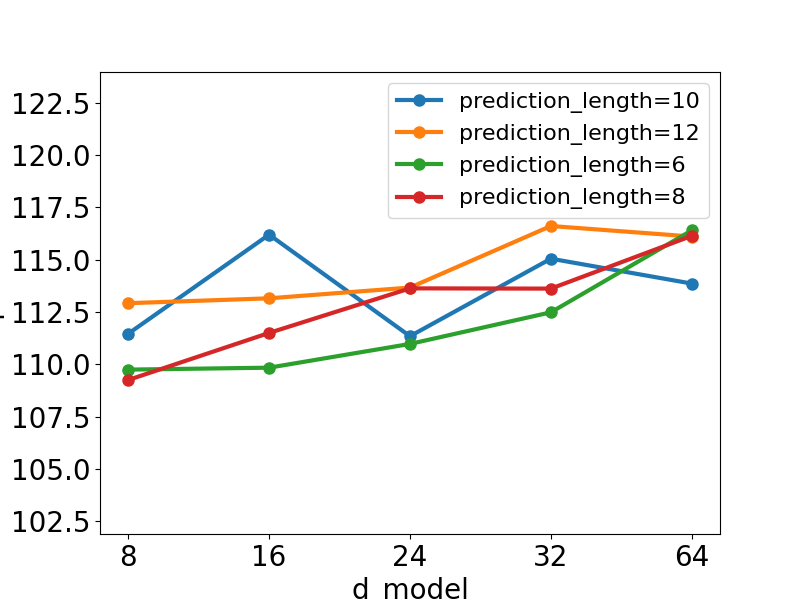}
}
\subfigure[Social Good]{
\includegraphics[width=0.31\textwidth]{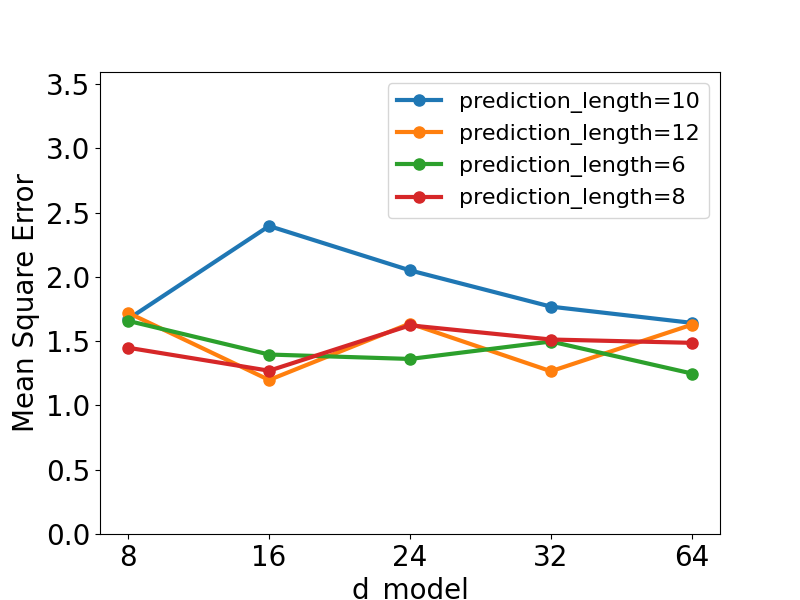}
}
\subfigure[Traffic]{
\includegraphics[width=0.31\textwidth]{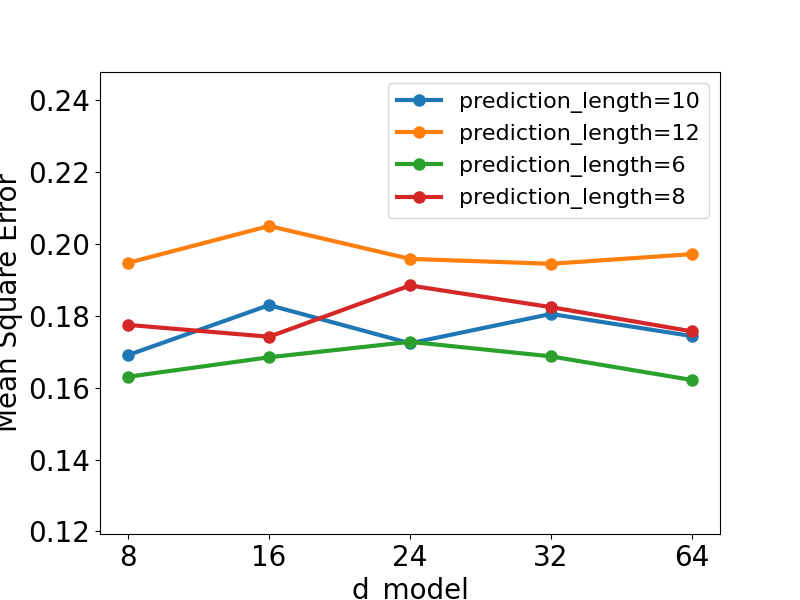}
}
\caption{Parameter study on model dimension $d$.}
\label{fig:param_study_dmodel}
\vspace{-3mm}
\end{figure*}

Figure~\ref{fig:param_study_dmodel} reveals several key insights about how model dimensionality affects forecasting performance. The plots demonstrate that increasing model dimensions generally improves accuracy (reduces MSE) but with domain-specific patterns and clear diminishing returns. While complex domains like Energy show substantial performance gains when scaling from small to medium dimensions (8→32), many other domains exhibit performance plateaus beyond certain thresholds, particularly for simpler or more periodic data patterns. Longer forecast horizons typically benefit more from increased dimensions than shorter ones, suggesting higher capacity requirements for extended predictions. The Security domain notably shows minimal dimension sensitivity, indicating that some forecasting challenges remain difficult regardless of model capacity. These patterns highlight the importance of balanced dimensionality selection—increasing capacity where complexity demands it, while avoiding unnecessary computation where simpler models suffice, enabling efficient resource allocation in practical deployments.

\begin{figure*}[h]
\centering
\vspace{-3mm}
\subfigure[Agriculture]{
\includegraphics[width=0.31\textwidth]{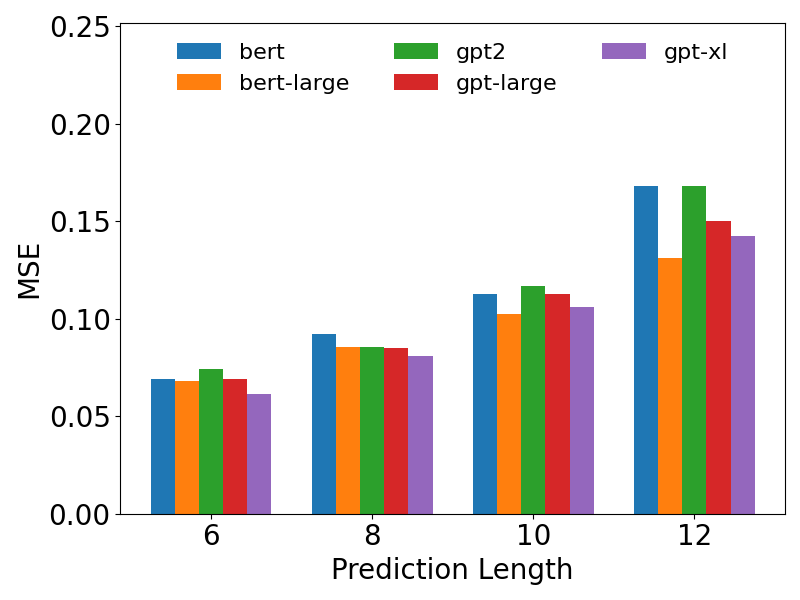}
}
\subfigure[Climate]{
\includegraphics[width=0.31\textwidth]{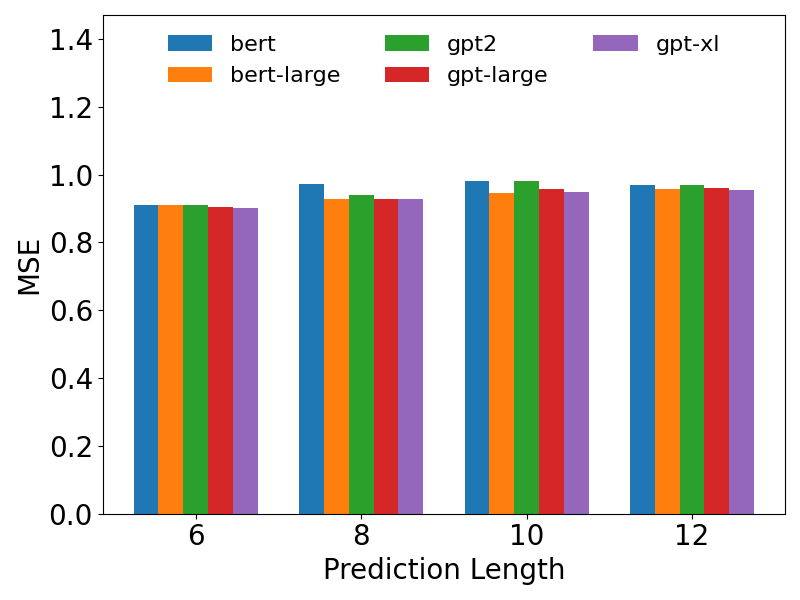}
}
\subfigure[Economy]{
\includegraphics[width=0.31\textwidth]{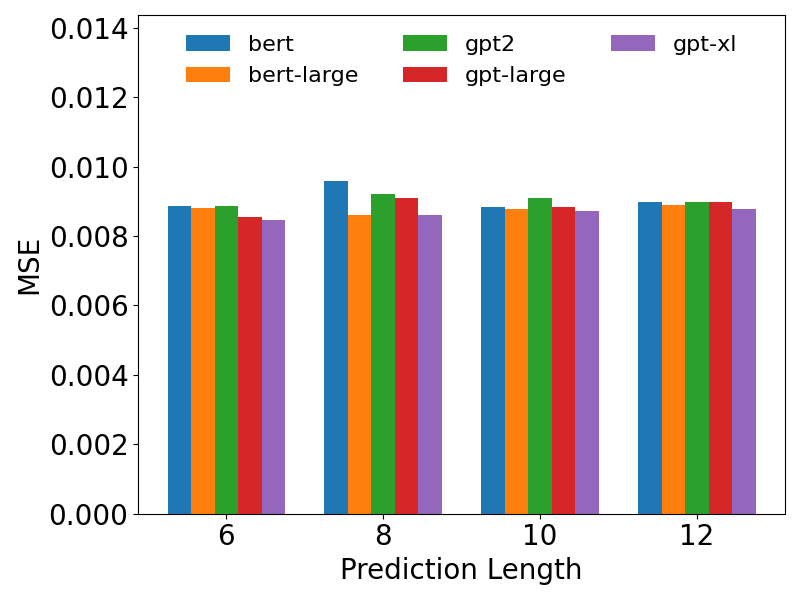}
}
\subfigure[Energy]{
\includegraphics[width=0.31\textwidth]{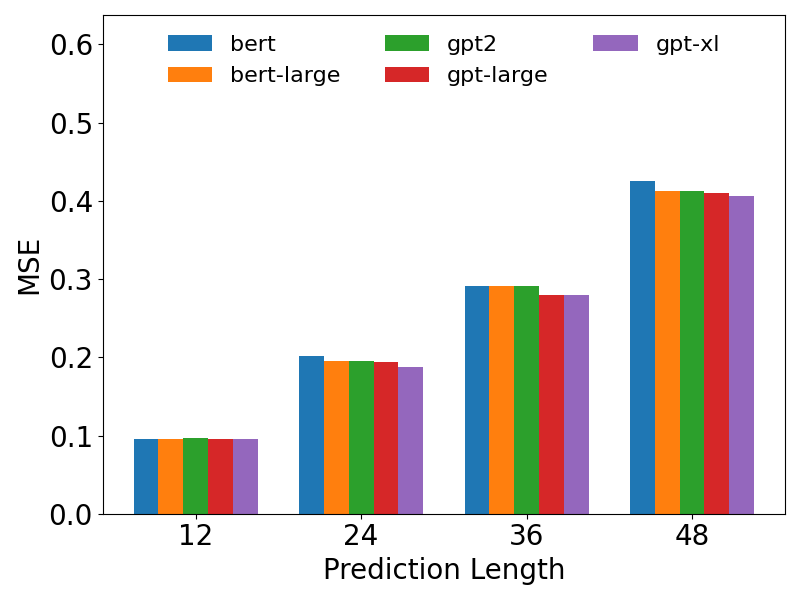}
}
\subfigure[Environment]{
\includegraphics[width=0.31\textwidth]{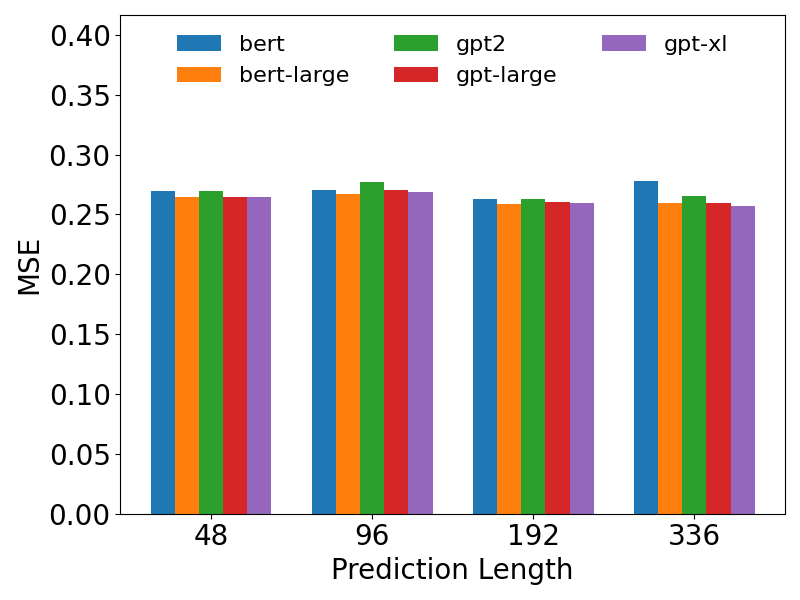}
}
\subfigure[Health]{
\includegraphics[width=0.31\textwidth]{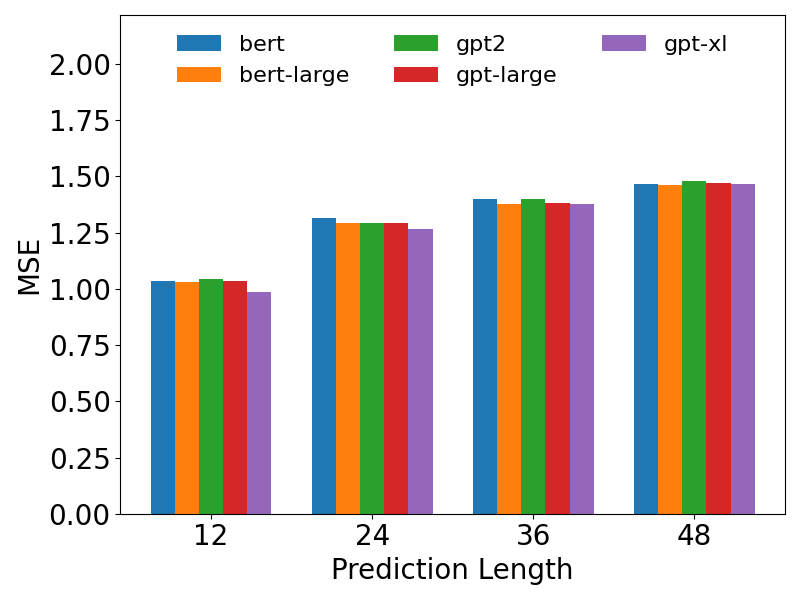}
}
\subfigure[Security]{
\includegraphics[width=0.31\textwidth]{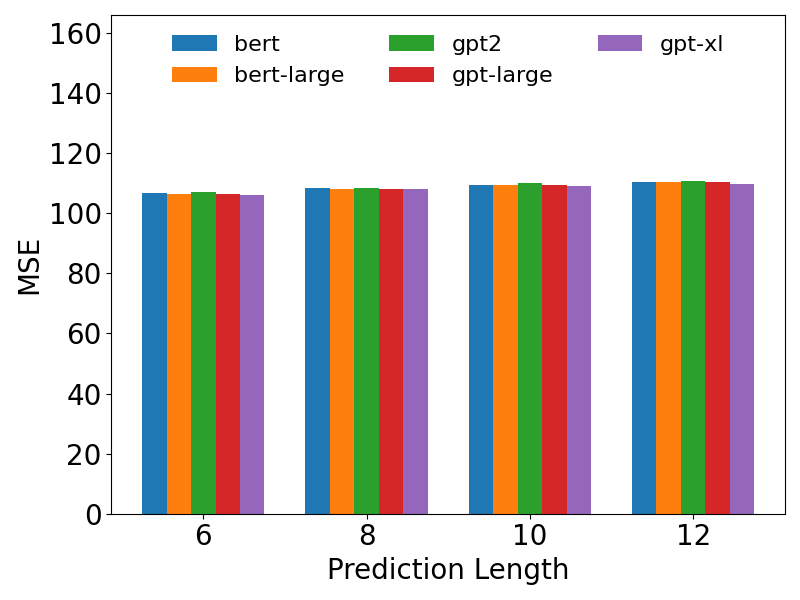}
}
\subfigure[Social Good]{
\includegraphics[width=0.31\textwidth]{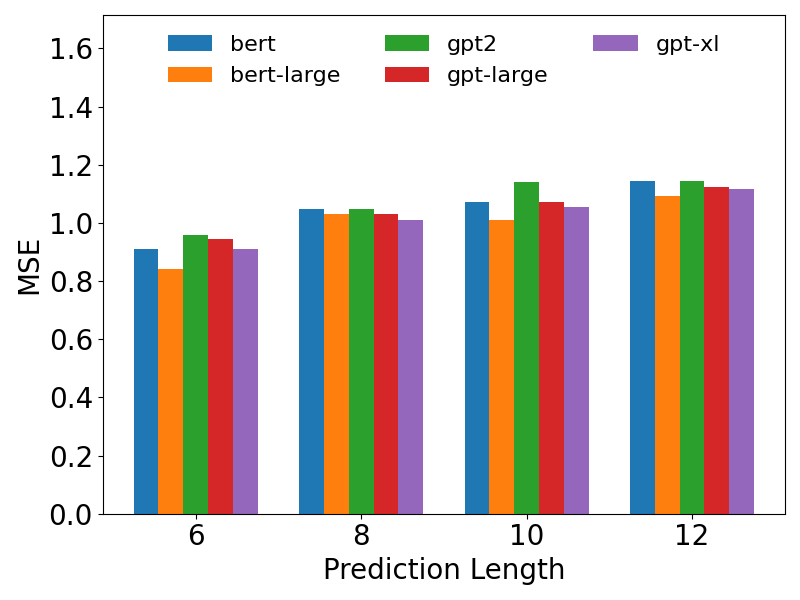}
}
\subfigure[Traffic]{
\includegraphics[width=0.31\textwidth]{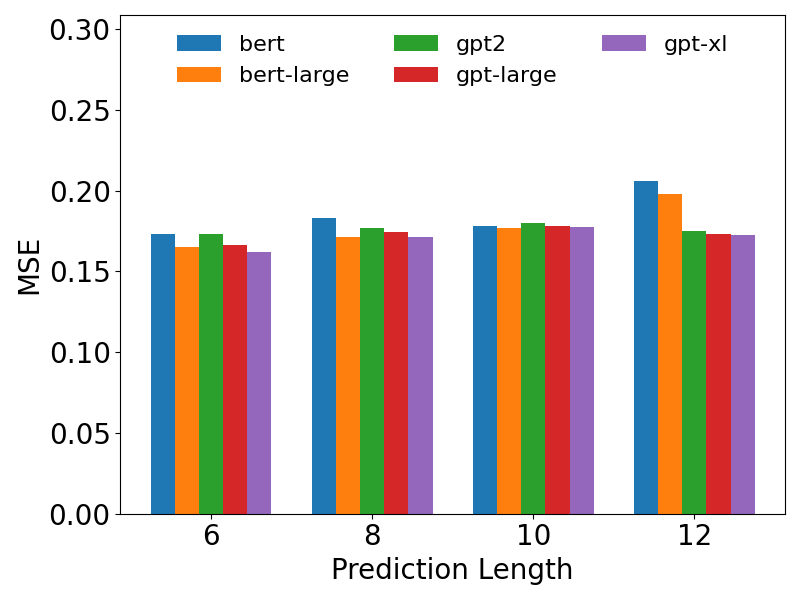}
}
\caption{Performance comparison of different language models.}
\label{fig:param_study_lm}
\vspace{-3mm}
\end{figure*}

Figure~\ref{fig:param_study_lm} demonstrates that increasing model scale consistently improves forecasting accuracy, with larger LMs (e.g., GPT-XL) achieving significantly lower MSE compared to smaller counterparts (e.g., BERT) in complex domains like Security (Figure A3g) and long-horizon predictions (e.g., 48-step Energy forecasts). This aligns with scaling laws in LLM research, where model capacity enhances pattern recognition in volatile or multi-scale data. However, smaller models (e.g., BERT-large) remain competitive in stable domains like Agriculture (Figure~\ref{fig:param_study_lm}a), achieving comparable accuracy with reduced computational costs. These results highlight a critical trade-off: practitioners should prioritize larger LLMs for high-stakes, irregular tasks while leveraging compact architectures for efficiency in predictable scenarios.

\clearpage

\clearpage

\clearpage

\section{Broader impacts}
\label{section:broader_impacts}
SpecTF framework advances multimodal forecasting accuracy while reducing computational costs, potentially benefiting applications like climate modeling and healthcare, where text-time series correlations drive critical decisions. However, ethical considerations arise regarding data privacy (e.g., sensitive textual inputs in medical records) and algorithmic bias propagation through spectral-text alignment, necessitating robust anonymization and fairness audits. The method’s efficiency (via RFFT) lowers energy consumption compared to dense time-domain models, aligning with sustainable AI practices. Yet, deployment in high-stakes domains like finance or security risks misuse for market manipulation or surveillance if not governed by transparency protocols. Future work should integrate explainability tools to audit text-frequency interactions and ensure equitable outcomes across socioeconomic groups.

\end{document}